\documentclass[10pt,twocolumn,letterpaper]{article}

\usepackage[pagenumbers]{cvpr} 

\usepackage{graphicx}
\usepackage{amsmath}
\usepackage{amssymb}
\usepackage{booktabs}
\usepackage{subcaption}
\usepackage{float}

\usepackage{hyperref}
\hypersetup{pagebackref,breaklinks,colorlinks}

\usepackage[capitalize]{cleveref}
\crefname{section}{Sec.}{Secs.}
\Crefname{section}{Section}{Sections}
\Crefname{table}{Table}{Tables}
\crefname{table}{Tab.}{Tabs.}

\usepackage{minitoc}
\usepackage{multirow}
\usepackage{mathtools}

\begin{document}

\newcommand{\vect}[1]{\ensuremath{\mathbf{#1}}}
\newcommand\todo[1]{\textcolor{red}{#1}}
\newcommand{\norm}[1]{\left\lVert#1\right\rVert}
\DeclarePairedDelimiterX{\infdivx}[2]{(}{)}{%
  #1\;\delimsize\|\;#2%
}
\newcommand{\zsem}{\vect{z}_\text{sem}}
\newcommand{\xT}{\vect{x}_T}
\newcommand{\xt}{\vect{x}_t}
\newcommand{\xzero}{\vect{x}_0}
\newcommand{\xtone}{\vect{x}_{t-1}}
\newcommand{\Ndist}{\mathcal{N}(\vect{0}, \mathbf{I})}

\title{Diffusion Autoencoders: Toward a Meaningful and Decodable Representation}

\newcommand{\mquad}{\hspace{0.5cm}}
\author{Konpat Preechakul \mquad Nattanat Chatthee    \mquad Suttisak Wizadwongsa \mquad Supasorn Suwajanakorn \\
VISTEC, Thailand\\
}

\twocolumn[{
\renewcommand\twocolumn[1][]{#1}
\maketitle
\vspace*{-10mm}

\begin{center}
    \label{mainmap}
    \centering
    \includegraphics[width=1\textwidth]{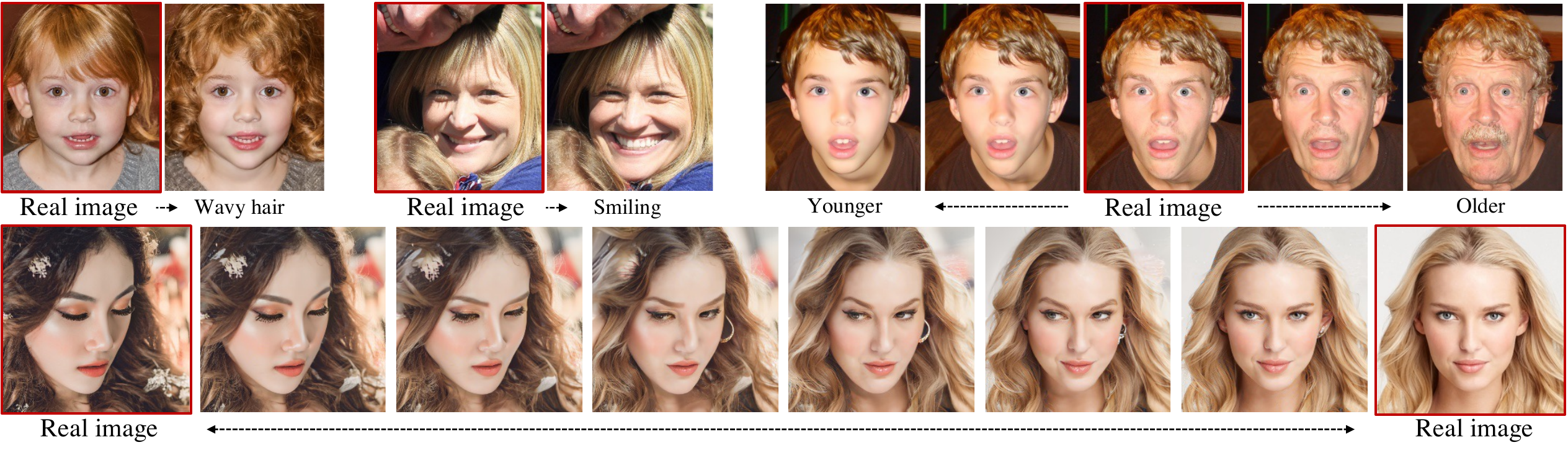}
    \vspace{-0.8cm}
    \captionof{figure}{\textbf{Attribute manipulation and interpolation on real images}. Diffusion autoencoders can encode any image into a two-part latent code that captures both semantics and stochastic variations and allows near-exact reconstruction. This latent code can be interpolated or modified by a simple linear operation and decoded back to a highly realistic output for various downstream tasks.
    }
    \label{fig:teaser}
\end{center}
}]

\begin{abstract}

Diffusion probabilistic models (DPMs) have achieved remarkable quality in image generation that rivals GANs'. But unlike GANs, DPMs use a set of latent variables that lack semantic meaning and cannot serve as a useful representation for other tasks. 
This paper explores the possibility of using DPMs for representation learning and seeks to extract a meaningful and decodable representation of an input image via autoencoding. Our key idea is to use a learnable encoder for discovering the high-level semantics, and a DPM as the decoder for modeling the remaining stochastic variations. 
Our method can encode any image into a two-part latent code where the first part is semantically meaningful and linear, and the second part captures stochastic details, allowing near-exact reconstruction. This capability enables challenging applications that currently foil GAN-based methods, such as attribute manipulation on \emph{real} images. We also show that this two-level encoding improves denoising efficiency and naturally facilitates various downstream tasks including few-shot conditional sampling. Please visit our page: \emph{\small \url{ https://Diff-AE.github.io/}}

\end{abstract}
\vspace{-0.4cm}
\section{Introduction}
\label{sec:intro}

Diffusion-based (DPMs) \cite{ho_denoising_2020, sohl-dickstein_deep_2015} and score-based \cite{song_score-based_2021} generative models 
have recently succeeded in synthesizing realistic and high-resolution images, rivaling those from GANs \cite{goodfellow_generative_2014, dhariwal_diffusion_2021, ho_cascaded_2022}. These two models are closely related and, in practice, optimize similar objectives. 
Numerous applications have emerged notably in the image domain, such as image manipulation, translation, super-resolution \cite{meng_sdedit_2021,saharia_image_2021,choi_ilvr_2021,li_srdiff_2022}, in speech and text domains \cite{austin_structured_2021,chen_wavegrad_2020}, or 3D point cloud\cite{luo_diffusion_2021}.  
Recent studies have improved DPMs further in both theory and practice \cite{kingma_variational_2021,lam_bilateral_2021,jolicoeur-martineau_gotta_2021}. 
In this paper, however, we question whether DPMs can serve as a good representation learner. Specifically, we seek to extract a meaningful and decodable representation of an image that contains high-level semantics yet allows near-exact reconstruction of the image. Our exploration focuses on diffusion models, but the contributions are applicable also to score-based models.

One way to learn a representation is through an autoencoder. 
There exists a certain kind of DPM \cite{song_denoising_2020} that can act as an encoder-decoder that converts any input image $x_0$ into a spatial latent variable $x_T$ by running the generative process backward. However, the resulting latent variable lacks high-level semantics and other desirable properties, such as disentanglement, compactness, or the ability to perform meaningful linear interpolation in the latent space. 
Alternatively, one can use a trained GAN for extracting a representation using the so-called GAN inversion \cite{karras_analyzing_2020,xia_gan_2021}, which optimizes for a latent code that reproduces the given input. Even though the resulting code carries rich semantics, this technique struggles to faithfully reconstruct the input image. To overcome these challenges, we propose a diffusion-based autoencoder that leverages the powerful DPMs for decodable representation learning.

Finding a meaningful representation that is decodable requires capturing both the high-level semantics and low-level stochastic variations. 
Our key idea is to learn both levels of representation by utilizing a learnable encoder for discovering high-level semantics and utilizing a DPM for decoding and modeling stochastic variations. 
In particular, we use our conditional variant of the Denoising Diffusion Implicit Model (DDIM) \cite{song_denoising_2020} as the decoder and separate the latent code into two subcodes. The first “semantic” subcode is compact and inferred with a CNN encoder, whereas the second “stochastic” subcode is inferred by reversing the generative process of our DDIM variant conditioned on the semantic subcode. 
In contrast to other DPMs, DDIM modifies the forward process to be non-Markovian while preserving the training objectives of DPMs. 
This modification allows deterministically encoding an image to its corresponding initial noise, which represents our stochastic subcode.

The implication of this framework is two-fold. First, by conditioning DDIM on the semantic information of the target output, denoising becomes easier and faster.
Second, this design produces a representation that is linear, semantically meaningful, and decodable---a novel property for DPMs' latent variables. This crucial property allows harnessing DPMs for many tasks including those that are highly challenging for any GAN-based methods, such as interpolation and attribute manipulation on \emph{real} images. Unlike GANs, which rely on 
error-prone inversion before operating on real images, our method requires no optimization to encode the input and produces high-quality output with original details preserved.

Despite being an autoencoder, which is generally not designed for unconditional generation, our framework can be used to generate image samples by fitting another DPM to 
the semantic subcode distribution.
This combination achieves competitive FID scores on unconditional generation compared to a vanilla DPM.  
Moreover, the ability to sample from our compact and meaningful latent space also enables few-shot conditional generation (i.e., generate images with similar semantics to those of a few examples). Compared to other DPM-based techniques for the few-shot setup, our method produces convincing results with only a handful labeled examples without additional contrastive learning used in prior work \cite{sinha_d2c_2021}.

\section{Background}

Diffusion-based (DPMs) and score-based generative models belong to a family of generative models that model the target distribution by learning a denoising process of varying noise levels.
A successful process can denoise or map an arbitrary Gaussian noise map from the prior $\mathcal{N}(\vect{0}, \mathbf{I})$ to a clean image sample after $T$ successive denoising passes.
Ho et al. \cite{ho_denoising_2020} proposed to learn a function $\epsilon_\theta(\xt, t)$ that takes a noisy image $\xt$ and predicts its noise using a UNet \cite{ronneberger_u-net_2015}. The model is trained with a loss function $\norm{\epsilon_\theta(\xt, t) - \epsilon}$, where $\epsilon$ is the actual noise added to $\xzero$ to produce $\xt$. This formulation is a simplified, reweighted version of the variational lower bound on the marginal log likelihood and has been commonly used throughout the community \cite{dhariwal_diffusion_2021,nichol_improved_2021,kingma_variational_2021,song_denoising_2020}. 

More formally, we define a Gaussian diffusion process at time $t$ (out of $T$) that increasingly adds noise to an input image $\xzero$ as $q(\xt|\xtone) = \mathcal{N}(\sqrt{1 - \beta_t}\xtone, \beta_t \mathbf{I})$, where $\beta_t$ are hyperparameters representing the noise levels. 
With Gaussian diffusion, the noisy version of an image $\xzero$ at time $t$ is another Gaussian $q(\xt|\xzero) = \mathcal{N}(\sqrt{\alpha_t}\xzero, (1 - \alpha_t) \vect{I})$ where $\alpha_t = \prod_{s=1}^t (1-\beta_s)$. 
We are interested in learning the reverse process of this, i.e., the distribution $p(\xtone|\xt)$. This probability function is likely a complex one unless the gap between $t-1$ and $t$ is infinitesimally small ($T = \infty$) \cite{sohl-dickstein_deep_2015}. In such a case, $p(\xtone|\xt)$ can be modeled as $\mathcal{N}(\mu_\theta(\xt, t), \sigma_t)$ \cite{ho_denoising_2020}. There are many ways to model this distribution, one of which is via $\epsilon_\theta(x_t, t)$ mentioned earlier. In practice, the assumption of $T = \infty$ is never satisfied; hence, DPMs are only approximations. 

As latent-variable models, DPMs can naturally yield the latent variables $\vect{x}_{1:T}$ through its forward process; however, these variables are stochastic and only representing a sequence of image degradation by Gaussian noise, which does not contain much semantics.
Song et al. \cite{song_denoising_2020} proposed another kind of DPM called Denoising Diffusion Implicit Model (DDIM) that enjoys the following deterministic generative process:
\begin{equation}
\resizebox{1\hsize}{!}{$\xtone = \sqrt{\alpha_{t-1}} \left( \frac{\xt - \sqrt{1 - \alpha_t} \epsilon_\theta^{t}(\xt)}{\sqrt{\alpha_t}} \right) + \sqrt{1 - \alpha_{t-1}} \epsilon_\theta^{t}(\xt)$}
\label{eq:gen}
\end{equation}
and the following novel inference distribution:
\begin{equation}
\resizebox{1\hsize}{!}{$
q(\xtone|\xt,\xzero) = \mathcal{N}\Big(\sqrt{\alpha_{t-1}} \xzero + \sqrt{1 - \alpha_{t-1}} \frac{\xt - \sqrt{\alpha_t} \xzero}{\sqrt{1 - \alpha_t}}, \vect{0} \Big)$}
\label{eq:ddimq}
\end{equation}
while maintaining the original DDPM marginal distribution $q(\xt|\xzero) = \mathcal{N}(\sqrt{\alpha_t}\xzero, (1-\alpha_t) \vect{I})$. By doing so, DDIM shares both the objective and solution with DDPM and only differs in how samples are generated.

With DDIM, it is possible to run the generative process backward deterministically to obtain the noise map $\xT$, which represents the latent variable or encoding of a given image $\xzero$. In this context, DDIM can be thought of as an image decoder that decodes the latent code $\xT$ back to the input image. This process can yield a very accurate reconstruction; however, $\xT$ still does not contain high-level semantics as would be expected from a meaningful representation. We show in Figure \ref{fig:interp_ddim} that the interpolation between two latent variables $\xT$'s does not correspond to a semantically-smooth change in the resulting images. The images only share the overall composition and background colors but do not resemble the identity of either person. This is, perhaps, understandable as $\xT$ is heavily influenced by the pixel values of $\xzero$ due to an implicit linear bias from the marginals $q(\xT|\xzero) = \mathcal{N}( \sqrt{\alpha_T}\xzero, (1 - \alpha_T )\mathbf{I})$. This motivates approaches that augment DPMs with novel mechanisms to make their latent variables more meaningful, as will be proposed in this work.

\section{Diffusion autoencoders}

\begin{figure}
  \centering
    \includegraphics[width=1\linewidth]{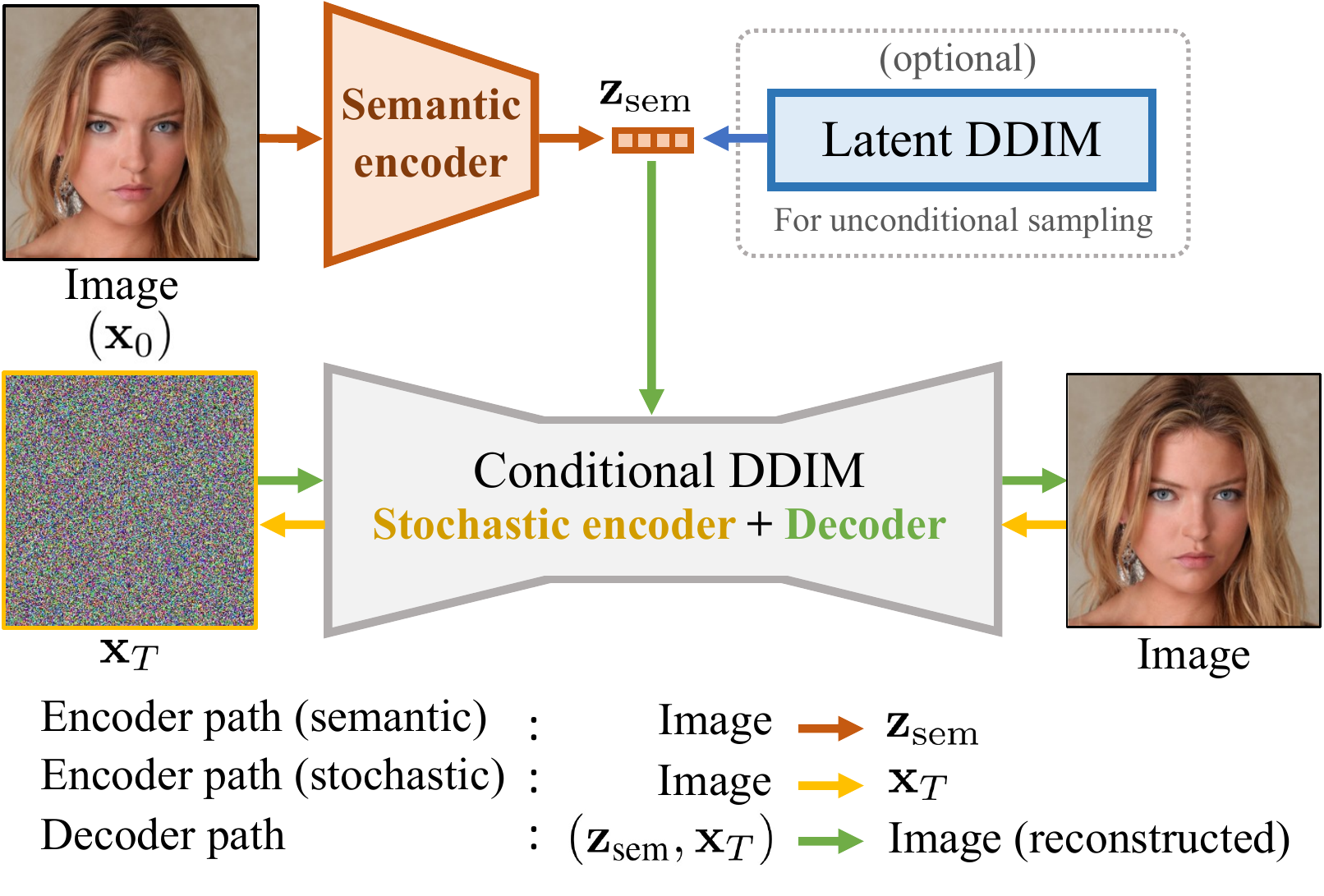}
    \vspace{-0.5cm}
    \caption{Overview of our diffusion autoencoder. The autoencoder consists of a ``semantic'' encoder that maps the input image to the semantic subcode $(\xzero \rightarrow \zsem)$, and a conditional DDIM that acts both as a ``stochastic'' encoder $(\xzero \rightarrow \xT)$ and a decoder $((\zsem, \xT) \rightarrow \xzero)$. Here, $\zsem$ captures the high-level semantics while $\xT$ captures low-level stochastic variations, and together they can be decoded back to the original image with high fidelity. To sample from this autoencoder, we fit a latent DDIM to the distribution of $\zsem$ and sample $(\zsem, \xT \sim \Ndist)$ for decoding.}
    \label{fig:framework}
\end{figure}
In the pursuit of a meaningful latent code, we design a conditional DDIM image \textbf{decoder} $p(\xtone|\xt, \zsem)$ that is conditioned on an additional latent variable $\zsem$, and a \textbf{semantic encoder} $\zsem=\text{Enc}_\phi(\xzero)$ that learns to map an input image $\xzero$ to a semantically meaningful $\zsem$.
Here, the conditional DDIM decoder takes as input a latent variable $\vect{z}=(\zsem, \xT)$, which consists of the high-level ``semantic'' subcode $\zsem$ and a low-level ``stochastic'' subcode $\xT$, inferred by reversing the generative process of DDIM. In this framework, DDIM acts as both the decoder and the stochastic encoder.
The overview is shown in Figure \ref{fig:framework}.

Unlike in other conditional DPMs \cite{sinha_d2c_2021, ho_cascaded_2022, li_srdiff_2022} that use spatial conditional variables (e.g., 2D latent maps), 
our $\vect{z}_\text{sem}$ is a non-spatial vector of dimension $d=512$, which resembles the style vector in StyleGAN \cite{karras_style-based_2019, karras_analyzing_2020} and allows us to encode global semantics not specific to any spatial regions.
One of our goals is to learn a semantically rich latent space that allows smooth interpolation, similar to those learned by GANs, while keeping the reconstruction capability that diffusion models excel.



\subsection{Diffusion-based Decoder}
\label{sec:deocder}
Our conditional DDIM decoder receives as input $\vect{z}=(\vect{z}_\text{sem}, \vect{x}_T)$ to produce the output image. 
This decoder is a conditional DDIM that models $p_\theta(\vect{x}_{t-1}|\vect{x}_t, \vect{z}_\text{sem})$ to match the inference distribution $q(\vect{x}_{t-1}|\vect{x}_t, \vect{x}_0)$ defined in Equation \ref{eq:ddimq}, with the following reverse (generative) process:
\begin{equation}
    p_\theta(\vect{x}_{0:T}\ |\ \zsem) = p(\vect{x}_T) \prod_{t=1}^Tp_\theta(\vect{x}_{t-1} \ |\ \vect{x}_t, \vect{z}_\text{sem})
\end{equation}
\begin{equation}
    p_\theta(\vect{x}_{t-1} | \xt, \zsem) = \begin{cases}
    \mathcal{N}(\vect{f}_\theta(\vect{x}_1, 1, \zsem), \vect{0}) & \hspace{-0.27cm}\text{if } t = 1 \\
    q(\vect{x}_{t-1}|\vect{x}_t, \vect{f}_\theta(\xt, t, \zsem)) & \hspace{-0.27cm}\text{otherwise}
  \end{cases}
    \label{eq:rev}
\end{equation}

Following Song et al. \cite{song_denoising_2020}, we parameterize $\vect{f}_\theta$ in Equation \ref{eq:rev} as a noise prediction network $\vect{\epsilon}_\theta(\vect{x}_t, t, \vect{z}_\text{sem})$:
\begin{equation}
\vect{f}_\theta(\xt, t, \zsem) = \frac{1}{\sqrt{\alpha_t}}\left( 
\xt - \sqrt{1 - \alpha_t} \vect{\epsilon}_\theta(\vect{x}_t, t, \vect{z}_\text{sem})
\right)
\label{eq:f_and_eps}
\end{equation}
This network is a modified version of the UNet of a recent DPM from Dhariwal et al.\cite{dhariwal_diffusion_2021}. Training is done by optimizing $L_\text{simple}$ \cite{ho_denoising_2020} loss function with respect to $\theta$ and $\phi$.
\begin{equation}
L_\text{simple} = \sum_{t=1}^T \mathbb{E}_{\xzero, \vect{\epsilon}_t} \Big[ 
\norm{\vect{\epsilon}_\theta(\vect{x}_t, t, \vect{z}_\text{sem}) - \vect{\epsilon}_t}_2^2
\Big]
\label{eq:loss}
\end{equation}
where $\vect{\epsilon}_t \in \mathbb{R}^{3 \times h \times w} \sim \mathcal{N}(\vect{0}, \vect{I})$, $\xt = \sqrt{\alpha_t}\xzero + \sqrt{1-\alpha_t} \vect{\epsilon}_t$, and $T$ is set to some large number, e.g., 1,000. Note that this simplified loss function has been shown to optimize both DDPM \cite{ho_denoising_2020} and DDIM \cite{song_denoising_2020}, though not the actual variational lower bound. For training, the stochastic subcode $\xT$ is not needed.
We condition the UNet using adaptive group normalization layers (AdaGN), following Dhariwal et al. \cite{dhariwal_diffusion_2021}, which extend group normalization~\cite{wu_group_2018} by applying channel-wise scaling and shifting on the normalized feature maps $\vect{h} \in \mathbb{R}^{c \times h \times w}$.
Our AdaGN is conditioned on $t$ and $\vect{z}_\text{sem}$: 
 \begin{equation}
 \text{AdaGN}(\vect{h}, t, \vect{z}_\text{sem}) = \vect{z}_s (\vect{t}_s \text{GroupNorm}(\vect{h}) + \vect{t}_b)
 \end{equation}
 where $\vect{z}_s \in \mathbb{R}^c = \text{Affine}(\vect{z}_\text{sem})$ and $(\vect{t}_s, \vect{t}_b) \in \mathbb{R}^{2 \times c} = \text{MLP}(\psi(t))$ is the output of a multilayer perceptron with a sinusoidal encoding function $\psi$.
These layers are used throughout the UNet. Please see details in Appendix \ref{app:arch}. 

\subsection{Semantic encoder}
The goal of the semantic encoder $\text{Enc}(\xzero)$ is to summarize an input image into a descriptive vector $\zsem = \text{Enc}(\xzero)$  with necessary information to help the decoder $p_\theta(\vect{x}_{t-1}|\vect{x}_t, \vect{z}_\text{sem})$ denoise and predict the output image. 
We do not assume any particular architecture for this encoder; however, in our experiments, this encoder shares the same architecture as the first half of our UNet decoder. One benefit of conditioning DDIM with information-rich $\zsem$ is more efficient denoising process, which will be discussed further in Section \ref{sec:faster}.

\subsection{Stochastic encoder}
Besides decoding, our conditional DDIM can also be used to encode an input image $\xzero$ to the stochastic subcode $\xT$ by running its deterministic generative process backward (the reverse of Equation \ref{eq:gen}):
\begin{equation}
    \vect{x}_{t+1} = \sqrt{\alpha_{t+1}}\vect{f}_\theta(\xt, t, \zsem) +\sqrt{1-\alpha_{t+1}}\vect{\epsilon}_\theta(\vect{x}_t, t, \vect{z}_\text{sem})
\end{equation}
We can think of this process as a \textbf{stochastic encoder} because $\xT$ is encouraged to encode only the information left out by $\zsem$, which has a limited capacity for compressing stochastic details.
By utilizing both semantic and stochastic encoders, our autoencoder can capture an input image to the very last detail while also providing a high-level representation $\zsem$ for downstream tasks. Note that the stochastic encoder is not used during training (Equation \ref{eq:loss}) and is used to compute $\xT$ for tasks that require exact reconstruction or inversion, such as real-image manipulation. 

\section{Sampling with diffusion autoencoders}
By conditioning the decoder on $\zsem$, diffusion autoencoders are no longer generative models. So, to sample from our autoencoder, we need an additional mechanism to sample $\zsem \in \mathbb{R}^d$ from the latent distribution.
While VAE is an appealing choice for this task, balancing between retaining rich information in the latent code
and maintaining the sampling quality in VAE is hard \cite{van_den_oord_neural_2017,rybkin_simple_2021,sinha_d2c_2021,rosca_distribution_2019}.
GAN is another choice, though it complicates training stability, which is one main strength of DPMs.
Here, we choose to fit another DDIM,
called \textbf{latent DDIM} $p_\omega(\vect{z}_{\text{sem},t-1}|\vect{z}_{\text{sem},t})$, to the latent distribution of $\zsem = \text{Enc}_\phi(\xzero), \ \xzero \sim p(\xzero)$. 
Analogous to Equation \ref{eq:f_and_eps} and \ref{eq:loss}, training the latent DDIM is done by optimizing $L_\text{latent}$ with respect to $\omega$:
\begin{equation}
L_\text{latent} = \sum_{t=1}^T \mathbb{E}_{\zsem, \vect{\epsilon}_t} \Big[ 
\norm{\vect{\epsilon}_\omega(\vect{z}_{\text{sem},t}, t) - \vect{\epsilon}_t}_1
\Big]
\label{eq:loss_latent}
\end{equation}
where $\vect{\epsilon}_t \in \mathbb{R}^d \sim \mathcal{N}(\vect{0}, \vect{I})$
, $\vect{z}_{\text{sem},t} = \sqrt{\alpha_t}\zsem + \sqrt{1-\alpha_t} \vect{\epsilon}_t$, and $T$ is the same as in the DDIM image decoder. For $L_\text{latent}$, we empirically found that $L_1$ works better than $L_2$ loss. Unlike for 1D/2D images, there is no well-established DPM architecture for non-spatial data, but we have found that deep MLPs (10-20 layers) with skip connections perform reasonably well. The details are provided in Appendix \ref{app:latentddim}.


We first train the semantic encoder ($\phi$) and the image decoder ($\theta$) via Equation \ref{eq:loss} until convergence. Then, we train the latent DDIM ($\omega$) via Equation \ref{eq:loss_latent} with the semantic encoder fixed. In practice, the latent distribution modeled by the latent DDIM is first normalized to have zero mean and unit variance. Unconditional sampling from a diffusion autoencoder is thus done by sampling $\zsem$ from the latent DDIM and unnormalizing it, then sampling $\xT \sim \mathcal{N}(\vect{0}, \vect{I})$, and finally decoding $\vect{z} = (\zsem, \xT)$ using the decoder. 


Our choice of training the latent DDIM post-hoc has a few practical reasons. First, since training the latent DDIM takes only a fraction of the full training time, post-hoc training enables quick experiments on different latent DDIMs with the same diffusion autoencoder. Another reason is to keep $\zsem$ as expressive as possible by not imposing any constraints, such as the prior loss in VAE\cite{kingma_auto-encoding_2013}, that can compromise the quality of the latent variables. 


\section{Experiments}
We now turn to assessing the properties of our learned latent space and demonstrating new capabilities, such as attribute manipulation and conditional generation. 
For fair comparison, the DDIM baseline in our experiments refers to our reimplementation of DDIM \cite{song_denoising_2020} based on an improved architecture of Dhariwal et al. \cite{dhariwal_diffusion_2021} with the same UNet hyperparameters as our decoder. In short, the DDIM baseline is similar to our decoder except that it does not take $\zsem$.
\subsection{Latent code captures both high-level semantics and low-level stochastic variations}
\begin{figure}
\begin{center}
    \centering
    \includegraphics[width=0.48\textwidth]{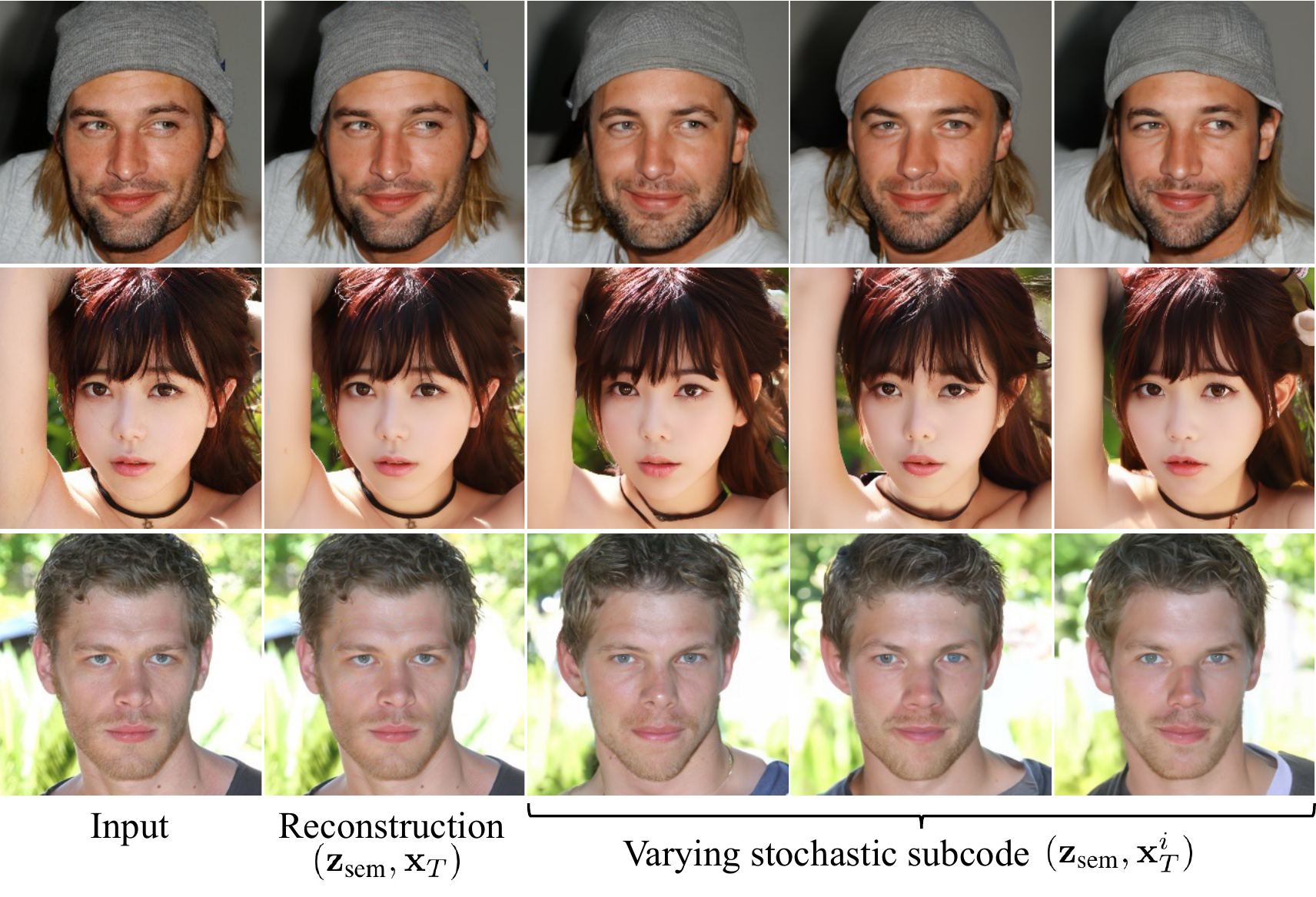}
    \vspace{-0.6cm}
    \caption{Reconstruction results and the variations induced by changing the stochastic subcode $\xT$. Each row corresponds to a different $\zsem$, which completely changes the person, whereas changing the stochastic subcode $\xT$ only affects minor details.
    }\label{fig:stoc}
\end{center}
\vspace{-0.5cm}
\end{figure}

To demonstrate that high-level semantics are mostly captured in $\zsem$ and very little in $\xT$, we first compute the semantic subcode $\zsem = \text{Enc}(\xzero)$ from an input image $\xzero$. For the stochastic subcode $\xT$, instead of inferring it from the input, we will sample this subcode multiple times $\xT^i \sim \mathcal{N}(\vect{0}, \mathbf{I})$ and decode multiple $\vect{z}^i = (\zsem, \xT^i)$. Figure \ref{fig:stoc} shows the variations induced by varying $\xT$ given the same $\zsem$, as well as the variations from different $\zsem$. 

The result show that with a fixed $\zsem$, the stochastic subcode $\xT$ only affects minor details, such as the hair and skin details, the eyes, or the mouth, but does not change the overall global appearance. And by varying $\zsem$, we obtain completely different people with different facial shapes, illuminations, and overall structures. Quantitative results are discussed in Section \ref{sec:autoencoding} and Table \ref{tab:recon}.

\subsection{Semantically meaningful latent interpolation}
\label{ex:interpolate}
\begin{figure}
  \centering
  \begin{subfigure}{1\linewidth}
    \includegraphics[width=1\textwidth]{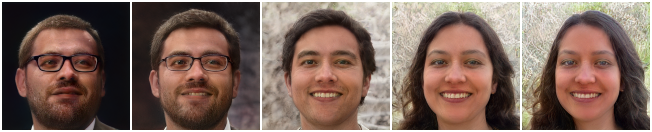}
    \vspace{-0.5cm}
    \caption{StyleGAN2 interpolation after $\mathcal{W}$ space inversion.}
    \vspace{-0.05cm}
    \label{fig:interp_w}
  \end{subfigure}
  \begin{subfigure}{1\linewidth}
    \includegraphics[width=1\textwidth]{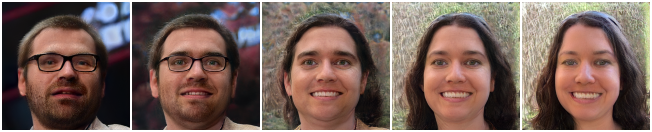}
    \vspace{-0.5cm}
    \caption{StyleGAN2 interpolation after $\mathcal{W}$+ space inversion.}
    \vspace{-0.05cm}
    \label{fig:interp_wp}
  \end{subfigure}
  \begin{subfigure}{1\linewidth}
    \includegraphics[width=1\textwidth]{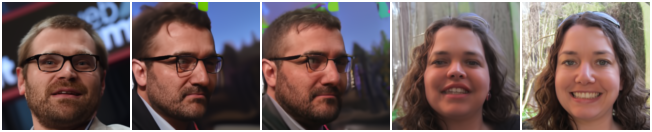}
    \vspace{-0.5cm}
    \caption{DDIM interpolation.}
    \vspace{-0.05cm}
    \label{fig:interp_ddim}
  \end{subfigure}
  \begin{subfigure}{1\linewidth}
    \includegraphics[width=1\textwidth]{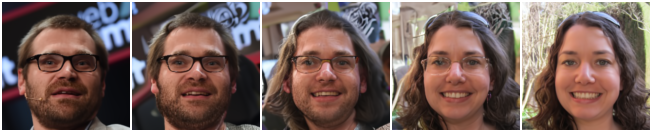}
    \vspace{-0.5cm}
    \caption{Our diffusion autoencoder interpolation.}
    \vspace{-0.05cm}
    \label{fig:interp_ours}
  \end{subfigure}
  \vspace{-0.2cm}
  \caption{Interpolation between two real images. In contrast to StyleGAN2 and DDIM, our method produces smooth and consistent results with well-preserved original details from both images.}
  \label{fig:interp}
  \vspace{-0.5cm}
\end{figure}

One desirable property of a useful latent space is the ability to represent semantic changes in the image by a simple linear change in the latent space. For example, by moving along a straight line connecting any two latent codes, we expect a smooth morphing between the corresponding two images. In Figure \ref{fig:interp_ours} and Figure \ref{fig:teaser}, we show our interpolation results by encoding two input images into $(\zsem^1, \xT^1)$ and $(\zsem^2, \xT^2)$, then decode $\vect{z}(t) = (\text{Lerp}(\zsem^1, \zsem^2; t), \text{Slerp}(\xT^1, \xT^2; t))$ for various values of $t \in [0, 1]$, where linear interpolation is used for $\zsem$ and spherical linear interpolation is used for $\xT$, following \cite{song_denoising_2020}. 

Compared to DDIM, which produces non-smooth transitions, our method gradually changes the head pose, background, and facial attributes between the two endpoints. The interpolation results from StyleGAN in both $\mathcal{W}$ and $\mathcal{W}+$ spaces are smooth, but the two endpoints do not resemble the input images, whereas ours and DDIM's match the real input images almost exactly. We quantitatively evaluate how smooth our interpolation is in Appendix \ref{app:ppl}.


\subsection{Attribute manipulation on real images}
\label{sec:attrib_man}
Another way to assess the relationship between image semantics and linear motion or separability in the latent space is by moving the latent $\zsem$ of an image in a particular direction and observing changes in the image\cite{shen_interpreting_2020}.
By finding such a direction from the weight vector of a linear classifier trained on latent codes of negative and positive images of a target attribute, e.g., smiling, this operation consequently changes the semantic attribute in the image.
There exists specialized techniques for this task \cite{abdal_styleflow_2021,shen_interpreting_2020, wu_stylespace_2021,patashnik_styleclip_2021}, but here we aim to showcase the quality and applicability of our latent space by using the simplest linear operation.


We trained linear classifiers using images and attribute labels from CelebA-HQ~\cite{karras_progressive_2018} and tested on CelebA-HQ and FFHQ~\cite{karras_style-based_2019} in Figure \ref{fig:attribute}. Implementation details and more results can be found in Appendix \ref{app:more_attr_manipulation}. Note that our autoencoder was trained on FFHQ but can generalize to CelebA-HQ without fine-tuning the autoencoder. 
Our method is able to change local features, such as the mouth for smiling, while keeping the rest of the image and details mostly stationary. For global attributes that involve changing multiple features at the same time, such as aging, our results look highly plausible and realistic. 
Additionally, we compare the accuracy of these linear classifiers (Appendix \ref{app:linear_classifier}) that take $\zsem$ versus those taking StyleGAN's inverted $\mathcal{W}$ as input. 
The AUROC$\uparrow$ over 40 attributes of our method is 0.925 and of StyleGAN-$\mathcal{W}$ is 0.891. 
And we test how much the input's identity is preserved via ArcFace~\cite{deng_arcface_2019} and quantify the manipulation quality in Appendix \ref{app:more_attr_manipulation}.

One notable advantage of diffusion autoencoders over GAN-based manipulation techniques is the ability to manipulate \emph{real} images while preserving details irrelevant to the manipulation (e.g., keeping the original hair and background when manipulating facial expression). When GANs are used for such tasks, the details are often altered because real images cannot be faithfully inverted back to the GAN's latent space. Compared to a recent score-based manipulation technique SDEdit~\cite{meng_sdedit_2021}, which focuses on local edits or translating images from another domain using a forward-backward sampling trick, our method solves changing semantic attributes by simply modifying the latent code. We also compare qualitatively to D2C~\cite{sinha_d2c_2021}, which uses NVAE\cite{vahdat_nvae_2020} decoder to perform a similar task in our Appendix \ref{app:manipulation}.

\begin{figure*}
\begin{center}
    \centering
    \includegraphics[width=1\textwidth]{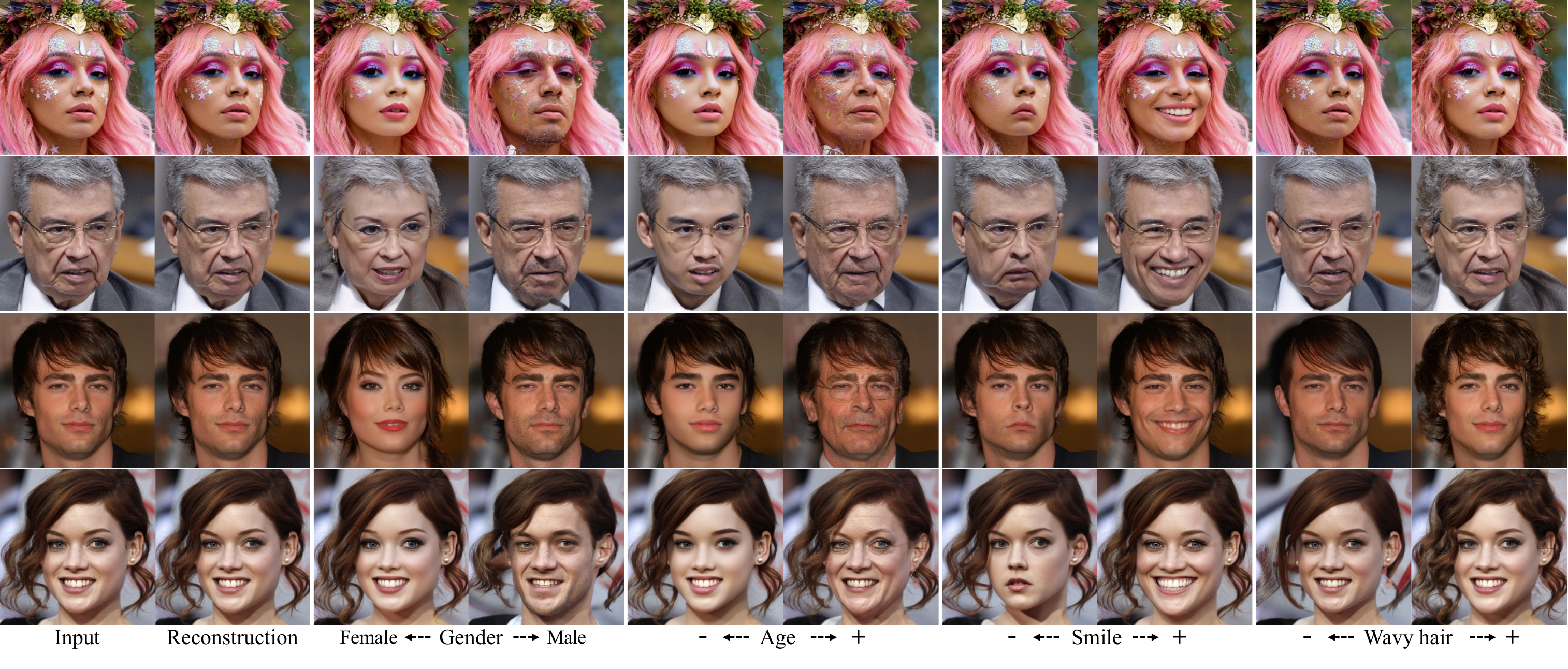}
    \vspace{-0.7cm}
    \caption{Real-image attribute manipulation results on two global attributes  (gender, age) and two local attributes (smile, wavy hair) by moving $\zsem$ along the positive or negative direction found by linear classifiers. The top two are from FFHQ~\cite{karras_style-based_2019} and the bottom two are from CelebA-HQ~\cite{karras_progressive_2018}. Our method synthesizes highly-plausible and realistic results that preserve an unprecedented level of detail.
    }\label{fig:attribute}
    \vspace{-0.5cm}
\end{center}
\end{figure*}

\subsection{Autoencoding reconstruction quality}
\label{sec:autoencoding}
Although good reconstruction quality of an autoencoder may not necessarily be an indicator of good representation learning, this property plays an important role in many applications, such as compression or image manipulation that requires accurate encoding-decoding abilities. For these tasks, traditional autoencoders that rely on MSE or $L_1$ loss functions perform poorly and produce blurry results. 
More advanced autoencoders combine perceptual loss and adversarial loss, e.g., VQGAN\cite{esser_taming_2021}, or rely on a hierarchy of latent variables, e.g., NVAE \cite{vahdat_nvae_2020}, VQ-VAE2 \cite{razavi_generating_2019}. Our diffusion autoencoder is an alternative design that produces a reasonable-size latent code with meaningful and compact semantic subcode and performs competitively with state-of-the-art autoencoders. The key is our two-level encoding that delegates the reconstruction of less compressible stochastic details to our conditional DDIM.

In Table \ref{tab:sota}, we evaluate the reconstruction quality of 1) our diffusion autoencoder, 2)  DDIM~\cite{song_denoising_2020}, 3) a pretrained StyleGAN2\cite{karras_analyzing_2020} (via two types of inversion), 4) VQ-GAN\cite{esser_taming_2021}, 5) VQ-VAE2\cite{razavi_generating_2019}, 6) NVAE\cite{vahdat_nvae_2020}. Both DDIM and ours were trained on 130M images and used T=100 for decoding. All these models were trained on FFHQ~\cite{karras_style-based_2019} and tested on 30k images from CelebA-HQ~\cite{karras_progressive_2018}. 
For our method and DDIM, we encoded downscaled test images of size 128$\times$128 and decoded them back. For the others, we used publicly available pretrained networks for 256$\times$256 and downscaled the results to the same 128$\times$128 before comparison. For StyleGAN2, we performed inversion in $\mathcal{W}$\cite{karras_analyzing_2020} and $\mathcal{W}+$\cite{abdal_image2stylegan_2019,abdal_image2stylegan_2020} spaces on the test images and used the optimized codes for reconstruction. 
The evaluation metrics are SSIM\cite{wang_image_2004} ($\uparrow$), LPIPS\cite{zhang_unreasonable_2018} ($\downarrow$), and MSE. NVAE\cite{vahdat_nvae_2020} achieves the lowest LPIPS and MSE scores, though it requires orders of magnitude larger latent dimension compared to others. Besides NVAE, our diffusion autoencoders outperform other models on all metrics, and only require $T$=20 steps to surpass DDIM with $T$=100 steps (Table \ref{tab:recon}). 
 
Furthermore, we performed ablation studies to investigate 1) the reconstruction quality when only $\zsem$ is encoded from the input but $\xT$ is sampled from $\Ndist$ for decoding (Table \ref{tab:recon}.a), and 2) the effects of varying the dimension of $\zsem$ from 64 to 512 (Table \ref{tab:recon}.b-e) on our autoencoder trained with 48M images for expedience. All configs a)-e) produce realistic results but differ in the degree of fidelity, where higher latent dimensions are better. For config a) with 512D $\zsem$, even though $\xT$ is random, the reconstructions still look perceptually close to the input images as measured by LPIPS (also Figure \ref{fig:stoc}). Our reconstruction with a small 64D $\zsem$ is already on par with StyleGAN2 inversion in 512D $\mathcal{W}$ latent space, suggesting that our diffusion autoencoders are proficient in compression.

\begin{table}[]
\caption{Autoencoding reconstruction quality of models trained on FFHQ~\cite{karras_style-based_2019} and tested on unseen CelebA-HQ~\cite{karras_progressive_2018}. Our model is competitive with state-of-the-art NVAE while producing readily useful high-level semantics in a compact 512D $\zsem$. 
}
\label{tab:sota}
\vspace{-1em}
\center
\footnotesize
\setlength{\tabcolsep}{2.5pt}
\begin{tabular}{lr|ccc}
\toprule
\textbf{Model}                  & \textbf{Latent dim} &  \textbf{SSIM} $\uparrow$ &\textbf{LPIPS} $\downarrow$ & \textbf{MSE} $\downarrow$ \\
\midrule
StyleGAN2 ($\mathcal{W}$) \cite{karras_analyzing_2020}  & 512   & 0.677  & 0.168  & 0.016     \\
StyleGAN2 ($\mathcal{W}$+) \cite{karras_analyzing_2020} & 7,168 & 0.827  & 0.114  & 0.006     \\
VQ-GAN           \cite{esser_taming_2021}             & 65,536    & 0.782  & 0.109  & 3.61e-3     \\
VQ-VAE2  \cite{razavi_generating_2019}              & 327,680   & 0.947  & 0.012  & 4.87e-4    \\
NVAE      \cite{vahdat_nvae_2020}                   &6,005,760  & 0.984  & \textbf{0.001}  & \textbf{4.85e-5}    \\
DDIM (T=100, $128^2$) \cite{song_denoising_2020}             & 49,152    & 0.917  & 0.063  & 0.002     \\
\textbf{Ours} (T=100, $128^2$, no $\xT$)                             & 512       & 0.677  & 0.073  & 0.007     \\
\textbf{Ours} (T=100, $128^2$)                                        & 49,664    & \textbf{0.991}  & 0.011  & 6.07e-5     \\
\bottomrule
\end{tabular}
\vspace{-1em}
\end{table}

\begin{table*}[]
\caption{
Ablation study results for a) autoencoding reconstruction quality when $\xT$ is not encoded from the input but sampled from $\Ndist$, and b-e) the effects of varying the dimension of $\zsem$ from 64 to 512 on our autoencoder trained with 48M images for expedience. In a), our reconstruction is perceptually close to the input images (LPIPS=0.073) even when $\xT$ is random. b-e) suggest that higher $\zsem$ dimensions lead to higher fidelity reconstruction. Our diffusion autoencoders with T=20 steps also surpass DDIM with T=100 steps.
}
\label{tab:recon}
\vspace{-1em}
\footnotesize
\begin{center}
\setlength{\tabcolsep}{5pt}
\begin{tabular}{l|cccc|cccc|cccc}
\toprule
\multirow{2}{*}{\textbf{Model}} 
& \multicolumn{4}{c|}{\textbf{SSIM} $\uparrow$} 
& \multicolumn{4}{c|}{\textbf{LPIPS} $\downarrow$}
& \multicolumn{4}{c}{\textbf{MSE} $\downarrow$} \\
& \multicolumn{1}{c}{\textbf{T=10}} 
& \multicolumn{1}{c}{\textbf{T=20}} 
& \multicolumn{1}{c}{\textbf{T=50}} 
& \multicolumn{1}{c|}{\textbf{T=100}} 
& \multicolumn{1}{c}{\textbf{T=10}}
& \multicolumn{1}{c}{\textbf{T=20}} 
& \multicolumn{1}{c}{\textbf{T=50}} 
& \multicolumn{1}{c|}{\textbf{T=100}} 
& \multicolumn{1}{c}{\textbf{T=10}} 
& \multicolumn{1}{c}{\textbf{T=20}} 
& \multicolumn{1}{c}{\textbf{T=50}} 
& \multicolumn{1}{c}{\textbf{T=100}} \\
\midrule
DDIM (@130M)\cite{song_denoising_2020} & 0.600 & 0.760 & 0.878 & 0.917 & 0.227 & 0.148 & 0.087 & 0.063 & 0.019 & 0.008 & 0.003 & 0.002 \\
\hline
\textbf{Ours} (@130M, 512D $\zsem$)            & 0.827 & 0.927 & 0.978 & \textbf{0.991} & 0.078 & 0.050 & 0.023 & \textbf{0.011} & 0.001 & 0.001 & 0.000 & \textbf{0.000} \\
a) No encoded $\xT$          & 0.707 & 0.695 & 0.683 & 0.677 & 0.085 & 0.078 & 0.074 & 0.073 & 0.006 & 0.007 & 0.007 & 0.007 \\

b) No encoded $\xT$, @48M, 512D $\zsem$ & 0.662  & 0.650 & 0.637 & 0.631 & 0.102 & 0.096 & 0.093 & 0.092 & 0.009 & 0.009 & 0.009 & 0.010 \\
c) No encoded $\xT$, @48M, 256D $\zsem$        & 0.637 & 0.624 & 0.612 & 0.606 & 0.116 & 0.109 & 0.106 & 0.105 & 0.010 & 0.011 & 0.011 & 0.011 \\
d) No encoded $\xT$, @48M, 128D $\zsem$        & 0.613 & 0.600 & 0.588 & 0.582 & 0.133 & 0.127 & 0.125 & 0.124 & 0.012 & 0.012 & 0.013 & 0.013 \\
e) No encoded $\xT$, @48M, 64D $\zsem$        & 0.551 & 0.538 & 0.527 & 0.521 & 0.168 & 0.165 & 0.163 & 0.162 & 0.018 & 0.019 & 0.020 & 0.020 \\

\bottomrule
\end{tabular}
\end{center}
\vspace{-0.5cm}
\end{table*}

\subsection{Faster denoising process}
\label{sec:faster}
One useful benefit of conditioning the denoising process with semantic information from $\zsem$ is faster generation. One main reason DPMs require many generative steps is because DPMs can only use a Gaussian distribution to approximate $p(x_{t-1}|x_t)$ when $T$ is sufficiently large ($\sim$1000).
Recent attempts to improve sampling speed focus on finding a better sampling interval or noise schedule~\cite{kingma_variational_2021,lam_bilateral_2021,jolicoeur-martineau_gotta_2021,nichol_improved_2021}, or using more efficient solvers to solve the score-based ODE counterpart \cite{jolicoeur-martineau_gotta_2021}.
Our diffusion autoencoders do not aim to solve this problem directly, nor can they be compared in the same context as generative models that lack access to the target samples. It is, however, worth mentioning the effects they have within the DPM framework. 

\begin{figure}
  \centering
  \begin{subfigure}{1\linewidth}
    \includegraphics[width=1\textwidth]{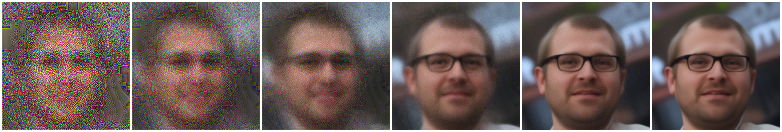}
    \vspace{-0.5cm}
    \caption{DDIM predicting $x_0$.}
    \vspace{-0.05cm}
  \end{subfigure}
  \vfill
  \begin{subfigure}{1\linewidth}
    \includegraphics[width=1\textwidth]{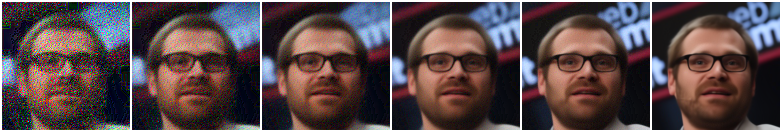}
    \vspace{-0.5cm}
    \caption{Our diffusion autoencoder predicting $x_0$.}
    \vspace{-0.05cm}
  \end{subfigure}
  \vspace{-0.5em}
  \caption{Predicted $\xzero$ at $t_{9, 8, 7, 5, 2, 0}$ ($T$=10). By conditioning on $\zsem$, our method predicts images that resemble $\xzero$ much faster.}
  \label{fig:gen_predx0}
  \vspace{-0.3cm}
\end{figure}

Consider a scenario where $\xzero$ is known to the denoising network. The noise prediction task will become trivial, and $q(\xtone|\xt, \xzero)$ is a Gaussian distribution regardless of the number of timesteps~\cite{ho_denoising_2020}. Since our diffusion autoencoders model the distribution $p(\xtone|\xt, \zsem)$, it follows that $p(\xtone|\xt, \zsem)$ is a better approximation to $q(\xtone|\xt, \xzero)$ than $p(\xtone|\xt)$ when $\zsem$ has captured much information about $\xzero$.
Figure \ref{fig:gen_predx0} shows that a diffusion autoencoder is able to predict $\xzero$ more accurately in fewer steps than DDIM and yield better image quality on four different datasets with the same timesteps $T$ in Table \ref{tab:uncond}. 

\subsection{Class-conditional sampling}
\label{sec:cond_sampling}
This experiment demonstrates how our framework can be used for few-shot conditional generation and compares to D2C~\cite{sinha_d2c_2021}, a state-of-the-art DPM-based method for this setup. We follow the problem setup in D2C where the goal is to generate a diverse set of images of a target class, such as female, by utilizing a small number of labeled examples ($\leq 100$). The labels can specify both the positives and negatives with respect to the target class (binary scenario) or only the positives (positive and unlabeled, or PU scenario). 
Given a latent classifier $p_\gamma(c|\zsem)$ for a target class $c$, one simple way to do class-conditional sampling is with rejection sampling, as used by D2C. That is, we sample $\zsem$ from our latent DDIM and accept this sample with probability $p_\gamma(c|\zsem)$. We followed D2C's methodology and conditionally sampled 5k images, then computed FID scores between these images and all the images of the same target class in CelebA dataset (with the same crop used by D2C). We used $T=100$ for both latent and image generations. Table \ref{tab:cond_vs_d2c} shows that our method achieves comparable FID scores to D2C, despite not using any self-supervised contrastive learning used in D2C.


\begin{table}[]
\caption{FID scores ($\downarrow$) for class-conditional generation on CelebA 64 dataset computed between 5k sampled images and the target subset. $\pm$ represents one standard deviation ($n$=3). D2C \cite{sinha_d2c_2021} results come from their paper ($n$=1 run of FID computation on 5k samples). Binary classifier was trained with 50 positives and 50 negatives. Positive-unlabeled (PU) classifier was trained with 100 positives and 10,000 unlabeled examples (as negatives). Naive FIDs were computed between all images and the target subset.
}
\label{tab:cond_vs_d2c}
\vspace{-1em}
\center
\footnotesize
\begin{tabular}{l|l|ccc}
\toprule
\textbf{Scenario}                              & \textbf{Classes}   & \textbf{Ours} & \textbf{D2C \cite{sinha_d2c_2021}}   & \textbf{Naive} \\
\midrule
\multirow{4}{*}{Binary}             & Male      & \textbf{11.52} $\pm$ 1.19               & 13.44 & 23.83 \\
                                    & Female    & \textbf{7.29} $\pm$ 0.44                 & 9.51  & 13.64 \\
                                    & Blond     & \textbf{16.10} $\pm$ 2.00                & 17.61 & 25.62      \\
                                    & Non-Blond & \textbf{8.48} $\pm$ 0.52                 & 8.94  & 0.96       \\
\hline
\multirow{4}{*}{PU} & Male      & \textbf{9.54} $\pm$ 0.54                 & 16.39 &     23.83  \\
                                    & Female    & \textbf{9.21} $\pm$ 0.19                & 12.21 & 13.64      \\
                                    & Blond     & \textbf{7.01} $\pm$ 0.25                 & 10.09 & 25.62      \\
                                    & Non-Blond & \textbf{7.91} $\pm$ 0.15                 & 9.09  & 0.96 \\
\bottomrule
\end{tabular}
\vspace{-0.3cm}
\end{table}

\subsection{Unconditional sampling}
\label{sec:uncond}
To evaluate the quality of our unconditional samples from diffusion autoencoders, we first sample $\zsem$ from the latent DDIM, then decode $\vect{z} = (\zsem, \xT \sim \mathcal{N}(\vect{0}, \vect{I}))$ using our decoder. We trained our autoencoders on FFHQ~\cite{karras_style-based_2019}, LSUN Horse \& Bedroom \cite{yu_lsun_2016}, and CelebA~\cite{liu_deep_2015}. For each dataset, we computed FID scores between 50k randomly sampled images from the dataset and our 50k generated images. We also varied the timestep $T$ = (10, 20, 50, 100) used in both latent DDIM and our main decoder.

As shown in Table \ref{tab:uncond}, our diffusion autoencoders are competitive with DDIM baselines and produce higher FID scores in most cases across numbers of timesteps. We also provide as reference our diffusion autoencoders trained with ground-truth latent variables encoded from the test images, labeled ``+autoencoding.'' In every dataset, perhaps unsurprisingly, conditioning the DDIM decoder with $\zsem$ significantly improves the quality with small $T$s. In Appendix \ref{app:not_memorize}, we show qualitative results and an additional experiment to verify that the latent DDIM does not memorize its input.

\begin{table}[]
\caption{FID scores ($\downarrow$) for unconditional generation. Our method is competitive with DDIM baselines. ``+ autoencoding'' refers to diffusion autoencoders that infer ground-truth semantic subcode from the test set and do not sample from the latent DDIM. 
}
\label{tab:uncond}
\vspace{-1.5em}
\center
\footnotesize
\setlength{\tabcolsep}{5pt}
\begin{tabular}{l|l|cccc}
\toprule
\multirow{2}{*}{\textbf{Dataset}} & \multirow{2}{*}{\textbf{Model}} & \multicolumn{4}{c} {\textbf{FID} $\downarrow$}            \\
                      &  & \textbf{T=10} & \textbf{T=20} & \textbf{T=50} & \textbf{T=100} \\
\midrule
FFHQ 128
& DDIM                   & 29.56  & 21.45  & 15.08  & 12.03   \\
&  \textbf{Ours}          & \textbf{20.80}  & \textbf{16.70}  & \textbf{12.57}  & \textbf{10.59} \\
&  + autoencoding         & 14.43  & 10.70  & 6.69   & 4.56    \\
\hline
Horse 128
& DDIM                   & 22.17  & 12.92  & 7.92   & \textbf{5.97}   \\
& \textbf{Ours}          & \textbf{11.97}  & \textbf{9.37}   & \textbf{7.44}   & 6.71    \\
& + autoencoding         & 9.27   & 6.23   & 3.87   & 2.92    \\
\hline
Bedroom 128
& DDIM                   & 13.70  & 9.23   & 7.14   & 5.94    \\
& \textbf{Ours}          & \textbf{10.69} & \textbf{8.19}   & \textbf{6.50}  & \textbf{5.70}  \\
& + autoencoding         & 6.36   & 4.88   & 3.61   & 2.88    \\
\hline
CelebA 64
& DDIM                   & 16.38  & 12.70  & 8.52   & 5.83    \\
& \textbf{Ours}          & \textbf{12.92}  & \textbf{10.18}   & \textbf{7.05}   & \textbf{5.30}    \\
& + autoencoding         & 12.78   & 9.06   & 5.15   & 3.11   \\
\bottomrule
\end{tabular}
\vspace{-1em}
\end{table}

\section{Related work}
Denoising diffusion-based generative models \cite{ho_denoising_2020,sohl-dickstein_deep_2015} are closely related to denoising score-based generative models \cite{song_generative_2019}. Models under this family have been shown to produce images with high quality rivaling those of GANs \cite{dhariwal_diffusion_2021} without using the less stable adversarial training. They are also used widely for multiple conditional generation tasks, such as image super-resolution \cite{saharia_image_2021,li_srdiff_2022}, image conditional generation \cite{meng_sdedit_2021,choi_ilvr_2021}, class-conditional generation in ImageNet dataset \cite{dhariwal_diffusion_2021}, and mel-spectrogram conditional speech synthesis \cite{chen_wavegrad_2020}. Similar to our work, these methods rely on conditional DPMs; however, most conditioning signals in prior work are known a priori and fixed, while our diffusion autoencoder augments the latent variable with an end-to-end learnable signal that the CNN encoder discovers. This puts our work closer to VAE\cite{kingma_auto-encoding_2013}, particularly Wehenkel et al. \cite{wehenkel_diffusion_2021} and D2C \cite{sinha_d2c_2021}. 
While these only utilize DPMs to model the prior distribution or latent representation for another generative model\cite{esser_imagebart_2021}, our focus is on how DPMs can be augmented with meaningful latent codes.

Our diffusion autoencoders share common goals with other kinds of autoencoders such as VAE \cite{kingma_auto-encoding_2013}, NVAE \cite{vahdat_nvae_2020}, and VQ-VAE \cite{van_den_oord_neural_2017} and VQ-VAE2 \cite{razavi_generating_2019}. While VAEs provide reasonable latent quality and sample quality, they are subject to posterior collapse \cite{van_den_oord_neural_2017} and prior holes problems \cite{sinha_d2c_2021} , whereas DPMs are not. VQ-VAE with discrete latent variables was proposed to deal with these problems by fitting an autoregressive Pixel-CNN model to the latent variable post-hoc \cite{van_den_oord_conditional_2016}. Fitting the latent variable post-hoc is also used in our work, but we utilize another DPM instead of an autoregressive model. 
Rich image representations are useful for many downstream tasks; for example, VAE are often used in model-based reinforcement learning \cite{ha_recurrent_2018,hafner_learning_2019,freeman_learning_2019} for predicting future outcomes of the environment. VQ-VAE's latent variables are used as a means for video generation tasks \cite{yan_videogpt_2021}. Our diffusion autoencoders also provide useful representations with an added ability to decode the representations back near perfectly.

Besides producing impressive image samples, GANs \cite{goodfellow_generative_2014} have been shown to learn meaningful latent spaces \cite{karras_style-based_2019} with extensive studies on multiple derived spaces \cite{harkonen_ganspace_2020,wu_stylespace_2021} and various knobs and controls for conditional human face generation \cite{he_attgan_2019,wu_relgan_2019,patashnik_styleclip_2021}.  
Encoding an image to the GAN's latent space requires an optimization-based inversion process \cite{karras_analyzing_2020,xia_gan_2021} or an external image encoder\cite{richardson_encoding_2021}, which has limited reconstruction fidelity (or yields high-dimensional codes outside the learned manifold).
This problem may be related to the GAN's limited latent size and mode-collapse problem, where the latent space only partially covers the support of training samples. Diffusion autoencoders do not have this problem and can readily encode any image without any additional error-prone optimization. 


\section{Limitations \& Discussion}
When encoding images that are out of the training distribution, our diffusion autoencoders can still reconstruct the images well, owing to the high-dimensional stochastic subcode from DDIM. 
However, both the inferred semantic and stochastic subcodes may fall outside the learned distributions, resulting in a poor representation that can no longer be interpreted or interpolated. While our choice of using non-spatial latent code is suitable for learning global semantics, certain image and spatial reasoning tasks may require more precise local latent variables. For these tasks, incorporating 2D latent maps can be beneficial.

For image generation, one unique feature of StyleGAN that is lacking from our diffusion autoencoders is the ability to control scale-specific generation. In terms of generation speed, our framework has significantly reduced the timesteps needed to achieve high-quality samples from our DDIM but still lacks behind GANs, which only require a single generator's pass to generate an image.

In conclusion, we have presented diffusion autoencoders that can separately infer both semantics and stochastic information from an input image. In contrast to DPMs and high-fidelity autoencoders like NVAE, our latent representation allows near-exact decoding while containing compact semantics readily useful for downstream tasks. These properties enable simple solutions to various \emph{real}-image editing tasks without requiring GANs and their error-prone inversion. Our framework also improves denoising efficiency and retains competitive unconditional sampling of DPMs.
{\small
\bibliographystyle{ieee_fullname}
\bibliography{bib}
}

\clearpage
\appendix 

\renewcommand \thepart{}
\renewcommand \partname{}
\addcontentsline{toc}{section}{Appendix} 
\part{Appendix} 
\section{Diffusion autoencoder architectures}
The baseline diffusion models and our diffusion autoencoders are based on the same DDIM model \cite{dhariwal_diffusion_2021} (publicly available at {\small \url{ https://github.com/openai/guided-diffusion}}). The architecture is specified in Table \ref{tab:arch_diffae}. We selected the hyperparameters differently due to the limited computational resources. 
Note that we used the linear $\beta$ scheduler as in Ho et al. \cite{ho_denoising_2020}, but we do observe improvements using the cosine $\beta$ scheduler \cite{nichol_improved_2021} in our preliminary results.
\label{app:arch}
\begin{figure}[h]
  \centering
  \begin{subfigure}{0.89\linewidth}
    \includegraphics[width=1\textwidth]{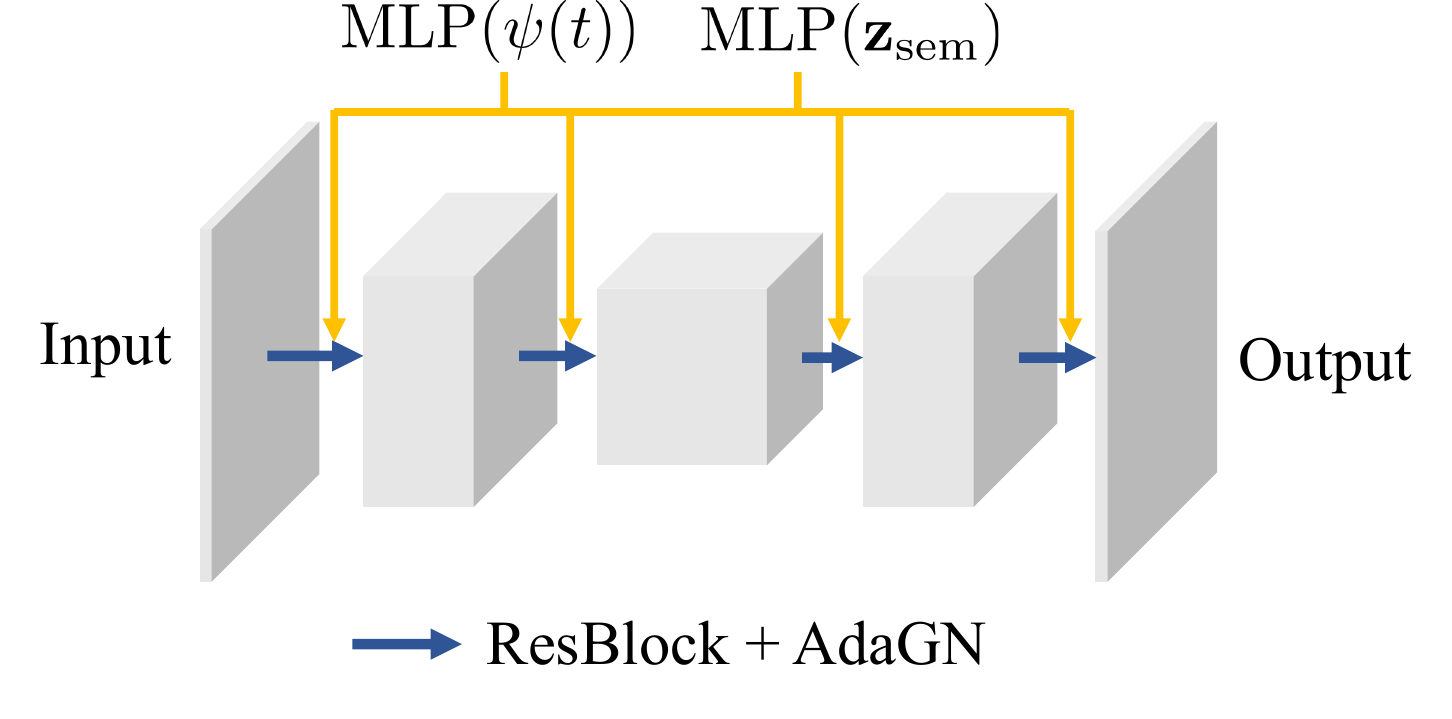}
    \caption{Diffusion autoencoder (Diff-AE)'s UNet decoder conditioned by $\zsem$.}
    \label{fig:arch1}
  \end{subfigure}
  \begin{subfigure}{0.85\linewidth}
    \includegraphics[width=1\textwidth]{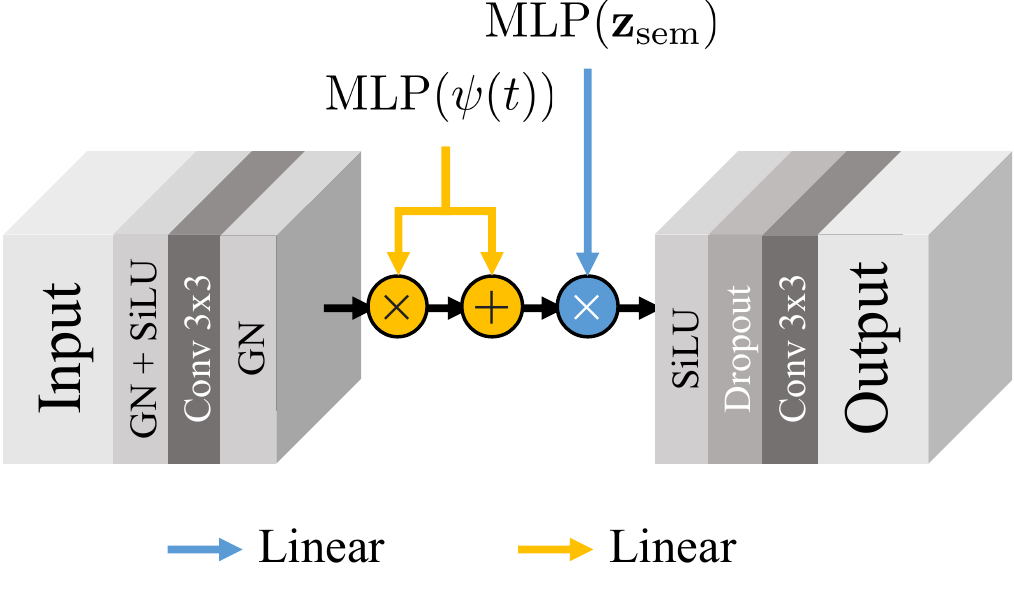}
    \caption{ResBlock + AdaGN. The residual path is not depicted.}
    \label{fig:arch2}
  \end{subfigure}
  \caption{Architecture overview of our diffusion autoencoder.}
  \label{fig:arch}
\end{figure}

\begin{table*}[t]
\caption{Network architecture of our diffusion autoencoder based on the improved DPM architecture of Dhariwal et al. \cite{dhariwal_diffusion_2021}.}
\label{tab:arch_diffae}
\vspace{-1em}
\begin{center}
\begin{tabular}{l|cccccc}
\toprule
\textbf{Parameter}           & \textbf{CelebA 64} & \textbf{FFHQ 64} & \textbf{FFHQ 128} & \textbf{Horse 128} & \textbf{Bedroom 128} & \textbf{FFHQ256} \\
\midrule
Batch size                   &  128 & 128              & 128               & 128                & 128   & 64               \\
Base channels      & 64 & 64           & 128               & 128                & 128         & 128         \\
Channel multipliers          & {[}1,2,4,8{]} &  {[}1,2,4,8{]}   & {[}1,1,2,3,4{]}   & {[}1,1,2,3,4{]}    & {[}1,1,2,3,4{]} & {[}1,1,2,2,4,4,{]}     \\
Attention resolution         & {[}16{]} & {[}16{]}        & {[}16{]}          & {[}16{]}           & {[}16{]} & {[}16{]}             \\
Images trained          & 72M & 48M              & 130M              & 130M                & 120M & 90M                  \\
Encoder base ch              & 64 & 64               & 128               & 128                & 128 & 128                 \\
Enc. attn. resolution & {[}16{]} & {[}16{]}         & {[}16{]}          & {[}16{]}           & {[}16{]} & {[}16{]}             \\
Encoder ch. mult.            & {[}1,2,4,8,8{]} & {[}1,2,4,8,8{]}  & {[}1,1,2,3,4,4{]} & {[}1,1,2,3,4,4{]}  & {[}1,1,2,3,4,4{]} & {[}1,1,2,2,4,4,4{]}   \\
$\zsem$ size                 & 512 & 512 & 512 & 512 & 512 & 512 \\
$\beta$ scheduler            & Linear & Linear & Linear & Linear & Linear & Linear \\
Learning rate                & \multicolumn{6}{c}{1e-4}                                                         \\
Optimizer                    & \multicolumn{6}{c}{Adam (no weight decay)}                                       \\
Training $T$                 & \multicolumn{6}{c}{1000}                                                         \\
Diffusion loss               & \multicolumn{6}{c}{MSE with noise prediction $\vect{\epsilon}$}                                    \\
Diffusion var.               & \multicolumn{6}{c}{Not important for DDIM} \\       
\bottomrule
\end{tabular}
\end{center}
\end{table*}

\subsection{Latent DDIM architectures}
\label{app:latentddim}
\begin{figure}[H]
\begin{center}
\includegraphics[width=0.38\textwidth]{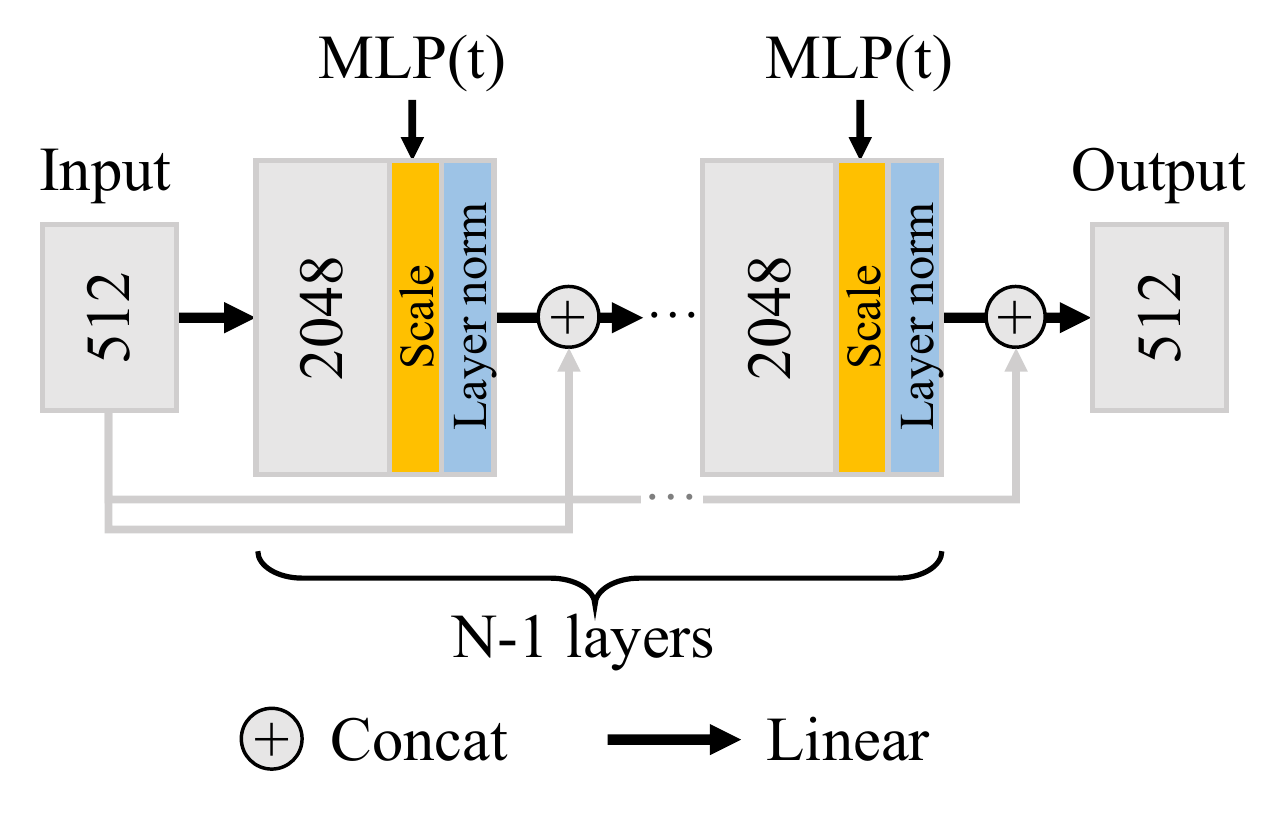}    
\caption{Architecture overview of our latent DDIM.}
\label{fig:archlatentddim}
\end{center}
\end{figure}

For latent DDIMs, we experimented with multiple architectures including MLP, MLP + skip connections, and projecting $\zsem$ into a spatial vector before using a CNN or UNet. We have found that MLP + skip connection performed reasonably well while being very fast (See unconditional samples in Figure \ref{fig:uncond_face}). The architecture is specified in Table \ref{tab:arch_latentddim}. Each layer of the MLP has a skip connection from the input, which simply concatenates the input with the output from the previous layer.
The network is conditioned on $t$ by scaling the hidden representations to help denoising. The architecture is shown in Figure \ref{fig:archlatentddim} and the hyperparameters are shown in Table \ref{tab:arch_latentddim}.

\begin{table*}
\caption{Network architecture of our latent DDIM.}
\label{tab:arch_latentddim}
\vspace{-1em}
\centering
\begin{tabular}{l|cccc}
\toprule
\textbf{Parameter}           & \textbf{CelebA} & \textbf{FFHQ} & \textbf{Horse} & \textbf{Bedroom} \\
\midrule
Batch size                   &  512 & 256 & 2048 & 2048           \\
$\zsem$ trained          & 300M & 100M & 2000M & 2000M                  \\
MLP layers ($N$)             & 10 & 10 & 20 & 20 \\
MLP hidden size                    & \multicolumn{4}{c}{2048} \\
$\zsem$ size                 & \multicolumn{4}{c}{512} \\
$\beta$ scheduler            & \multicolumn{4}{c}{Constant 0.008}  \\
Learning rate                & \multicolumn{4}{c}{1e-4}                                                         \\
Optimizer                    & \multicolumn{2}{c}{AdamW (weight decay = 0.01)} & \multicolumn{2}{c}{Adam (no weight decay)}                                       \\
Train Diff T                 & \multicolumn{4}{c}{1000}                                                         \\
Diffusion loss               & \multicolumn{4}{c}{L1 loss with noise prediction $\vect{\epsilon}$}                                    \\
Diffusion var.               & \multicolumn{4}{c}{Not important for DDIM} \\       
\bottomrule
\end{tabular}
\end{table*}
We have compared different $\beta$ schedulers including Linear \cite{ho_denoising_2020}, and a constant of 0.008 schedulers. (We found that Cosine \cite{nichol_improved_2021} scheduler underperformed during preliminary experiments for our latent DDIM.) We compared the two schedulers on the $\zsem$ of LSUN's Horse 128 diffusion autoencoder model. The latent DDIM is MLP + Skip with 10 layers and 2048 hidden nodes. The validation FID score for using linear beta schedule is 13.36, whereas for constant 0.008 scheduler is 10.50.
We found that an L1 loss performed better for the latent DDIM with FID of 11.65 vs 13.36 of MSE (Though, the main autoencoder uses MSE loss). 
We provide the hyperparameter tuning results of the MLP + Skip network: 

\begin{center}
\begin{tabular}{lr}
\toprule
\textbf{Latent model}         & \multicolumn{1}{l}{\textbf{FID}} \\
\midrule
Linear $\beta$, 10 layers, size 2048                 & 13.36                   \\
Constant 0.008 \& L1 & \multicolumn{1}{l}{}    \\
- 10 layers          & 10.16                   \\
\quad - size 3072          & 9.57                    \\
\quad - size 4096          & 9.43                    \\
- 15 layers          & 9.58                    \\
- 20 layers          & \textbf{9.30}                    \\
\bottomrule
\end{tabular}
\end{center}

Even though these results come from LSUN's Horse dataset, we found that similar settings worked well across datasets. We only tuned the network depth and the total training iterations for each dataset separately, a common practice in StyleGAN's training on these datasets.

\subsection{Classifiers}

We always use linear classifiers (logistic regression) trained on $\zsem$ space in all relevant experiments, which are attribute manipulation and class-conditional sampling. 
For training, $\zsem$ is first normalized so that its entire distribution has zero mean and unit variance before putting to the classifier. For the PU classifier, we oversampled the positive data points to match the negative ones to maintain the balance. For conditional generation, we follow D2C and apply rejection sampling after an additional thresholding. That is, we reject samples with the target class probabilities less than 0.5 before performing rejection sampling.

\section{Computation resources}
We used four Nvidia V100s for both diffusion autoencoders and DDIM and a single Nvidia RTX 2080 Ti for the latent DDIMs. Training the latent DDIMs takes only a fraction of the computational resources compared to the diffusion autencoders. Table \ref{tab:throughput} shows the throughputs of DDIM and diffusion autoencoders. Diffusion autoencoders were around 20\% slower to train than DDIM counterparts due to the additional semantic encoder. The total GPU-hours can be computed by multiplying the throughput with the number of training images for each model provided in Table \ref{tab:arch_latentddim}.
\begin{table}[H]
\caption{Throughputs of DDIM and diffusion autoencoders.}
\label{tab:throughput}
\vspace{-1em}
\begin{center}
\resizebox{\columnwidth}{!}{
\begin{tabular}{l|cc}
\toprule
\multirow{3}{*}{\textbf{Model}} & \textbf{DDIMs}   & \textbf{Diffusion autoencoders} \\
                      & Throughput & Throughput      \\
                     & (imgs/sec./V100) & (imgs/sec./V100) \\
\midrule
FFHQ-64                & 160        & 128             \\
FFHQ-128               & 51        & 41.65            \\
FFHQ-256               & -             & 10.08           \\
Horse-128              & 51           & 41.65                 \\
Bedroom-128            & 51       & 41.65             \\    
\bottomrule
\end{tabular}}
\end{center}
\end{table}

\section{Does the latent DDIM memorize its input?}
\label{app:not_memorize}
\begin{figure*}
 \centering
  \includegraphics[width=1\textwidth]{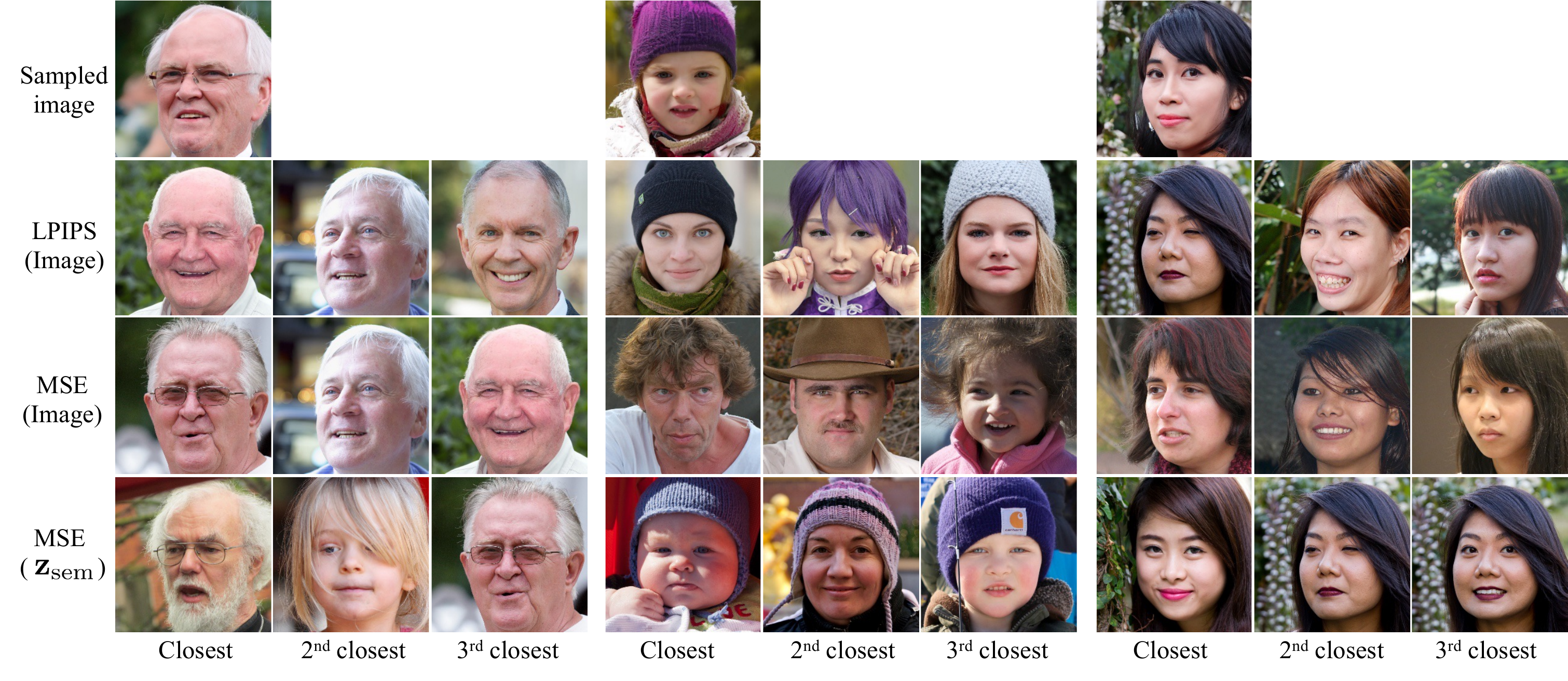}
  \caption{\textbf{Does latent DDIM memorize its input?} For each sampled image at the top, we find its closest images from the training set in terms of LPIPS, MSE in the image space, and MSE in the semantic subcode $\zsem$ space. The sampled images do not closely resemble any of the training images, suggesting that our latent DDIM does not memorize the input samples.}
  \label{fig:closest}
\end{figure*}
\begin{figure*}
 \centering
  \includegraphics[width=1\textwidth]{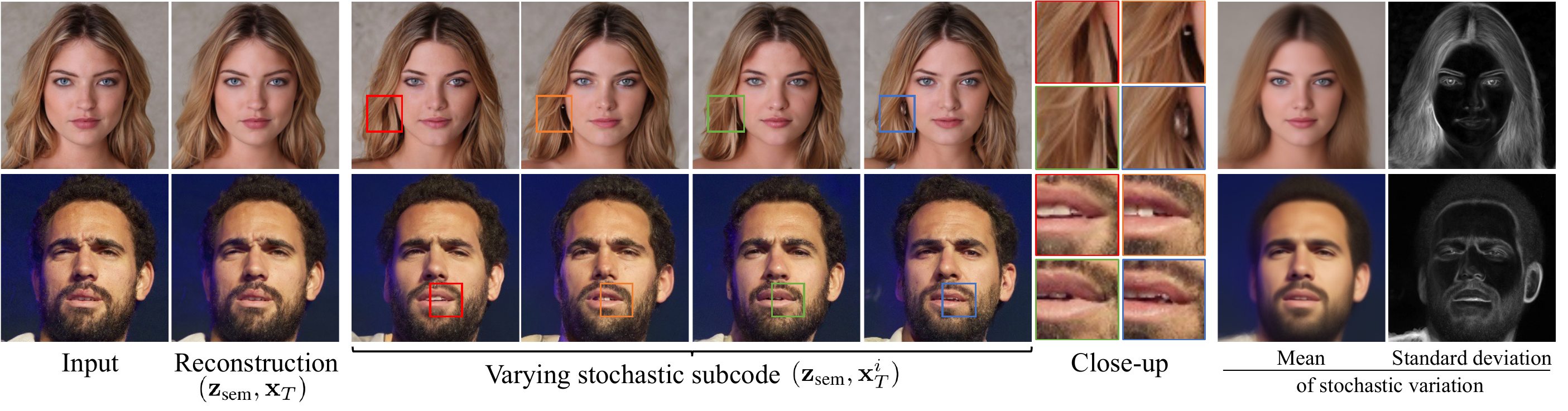}
  \caption{Reconstruction results and the variations induced by changing the stochastic subcode $\xT$.}
  \label{fig:stochastic_variation}
  \vspace{-0.05cm}
\end{figure*}
To verify if our diffusion autoencoder and latent DDIM can generate novel samples and do not simply memorize the input, we generate image samples and compare them to their nearest neighbors in the training set (Figure \ref{fig:closest}). (They should look different). To find nearest neighbors, we used three different metrics: 1) lowest LPIPS\cite{zhang_unreasonable_2018} in the image space, 2) lowest MSE in the image space, 3) lowest MSE in the semantic subspace $(\zsem)$. We have found that our autoencoder can generate substantially different images from the training set, suggesting no memorization problem. 

\section{What is encoded in the stochastic subcode?}
Figure \ref{fig:stochastic_variation} shows the stochastic variations induced by varying $\xT$ given the same $\zsem$. We also compute the mean and standard deviation of these variations. All generated images look realistic and $\xT$ changes only minor details, such as the hair pattern, while keeping the overall structure the same.

\section{Predictive power of the semantic subcode}
\label{app:linear_classifier}
We assess the quality of our proposed $\zsem$ via linear classification performance, which has been extensively used to evaluate the quality of learned representations \cite{alain_understanding_2018,chen_simple_2020,he_momentum_2020,grill_bootstrap_2020}. 
In Table \ref{tab:auroc}, we measure the performance of linear classifiers trained on $\zsem$ and StyleGAN's latent code in $\mathcal{W}$ space (obtained from an inversion process \cite{karras_analyzing_2020}) using Area Under the Receiver Operating Characteristic (AUROC) on the CelebA-HQ's 40 attributes with 30\% test data out of 30,000 total data points. The classifiers were trained on z-normalized latent vectors until convergence with Adam optimizer (learning rate 1e-3). For most classes, the linear classifiers using $\zsem$ outperform those using StyleGAN's $\mathcal{W}$ with weighted averages of 0.92 vs 0.89. This suggests that $\zsem$ contains attribute-specific information that is more readily predictive than that of StyleGAN's $\mathcal{W}$.

\begin{table}[h]
\caption{\label{tab:auroc} Classification AUROC $\uparrow$ on CelebA-HQ's 40 attributes of linear classifiers trained on our $\zsem$ vs. StyleGAN's latent code in $\mathcal{W}$ space (obtained via inversion).}
\vspace{-1em}
\center
\begin{tabular}{lrrr}
\toprule
                              \textbf{Class} & \multicolumn{1}{l}{\textbf{\#Positives}} & \multicolumn{1}{l}{$\zsem$} & \multicolumn{1}{l}{$\mathcal{W}$} \\
\midrule
5\_o\_Clock\_Shadow       & 1318                               & \textbf{0.96}                       & 0.94                                    \\
Arched\_Eyebrows          & 3262                               & \textbf{0.88}                       & 0.86                                    \\
Attractive                & 5183                               & \textbf{0.90}                       & 0.86                                    \\
Bags\_Under\_Eyes         & 2564                               & \textbf{0.89}                       & 0.85                                    \\
Bald                      & 229                                & 0.99                                & 0.99                                    \\
Bangs                     & 1601                               & \textbf{0.98}                       & 0.95                                    \\
Big\_Lips                 & 3247                               & \textbf{0.73}                       & 0.68                                    \\
Big\_Nose                 & 2813                               & \textbf{0.88}                       & 0.85                                    \\
Black\_Hair               & 1989                               & \textbf{0.96}                       & 0.93                                    \\
Blond\_Hair               & 1546                               & \textbf{0.99}                       & 0.97                                    \\
Blurry                    & 34                                 & \textbf{0.90}                       & 0.82                                    \\
Brown\_Hair               & 2087                               & \textbf{0.89}                       & 0.81                                    \\
Bushy\_Eyebrows           & 1682                               & \textbf{0.93}                       & 0.85                                    \\
Chubby                    & 622                                & \textbf{0.95}                       & 0.93                                    \\
Double\_Chin              & 530                                & \textbf{0.95}                       & 0.94                                    \\
Eyeglasses                & 416                                & 1.00                                & 0.98                                    \\
Goatee                    & 688                                & \textbf{0.98}                       & 0.96                                    \\
Gray\_Hair                & 395                                & \textbf{0.98}                       & 0.97                                    \\
Heavy\_Makeup             & 4143                               & \textbf{0.97}                       & 0.95                                    \\
High\_Cheekbones          & 4160                               & \textbf{0.95}                       & 0.91                                    \\
Male                      & 3273                               & 1.00                                & 1.00                                    \\
Mouth\_Slightly\_Open     & 4195                               & \textbf{0.98}                       & 0.94                                    \\
Mustache                  & 502                                & \textbf{0.97}                       & 0.94                                    \\
Narrow\_Eyes              & 998                                & \textbf{0.86}                       & 0.77                                    \\
No\_Beard                 & 7335                               & \textbf{0.99}                       & 0.97                                    \\
Oval\_Face                & 1872                               & \textbf{0.77}                       & 0.71                                    \\
Pale\_Skin                & 434                                & \textbf{0.96}                       & 0.94                                    \\
Pointy\_Nose              & 2855                               & \textbf{0.74}                       & 0.70                                    \\
Receding\_Hairline        & 777                                & \textbf{0.94}                       & 0.89                                    \\
Rosy\_Cheeks              & 1003                               & \textbf{0.96}                       & 0.92                                    \\
Sideburns                 & 747                                & \textbf{0.99}                       & 0.97                                    \\
Smiling                   & 4175                               & \textbf{0.99}                       & 0.96                                    \\
Straight\_Hair            & 1975                               & \textbf{0.84}                       & 0.77                                    \\
Wavy\_Hair                & 3197                               & \textbf{0.90}                       & 0.87                                    \\
Wearing\_Earrings         & 2310                               & \textbf{0.92}                       & 0.86                                    \\
Wearing\_Hat              & 325                                & \textbf{0.99}                       & 0.96                                    \\
Wearing\_Lipstick         & 5064                               & \textbf{0.98}                       & 0.97                                    \\
Wearing\_Necklace         & 1501                               & \textbf{0.79}                       & 0.75                                    \\
Wearing\_Necktie          & 636                                & \textbf{0.96}                       & 0.95                                    \\
Young                     & 6978                               & \textbf{0.94}                       & 0.91                                    \\
\textbf{Weighted average} & \multicolumn{1}{l}{}               & \textbf{0.92}                       & 0.89                                    \\
\textbf{Macro average}          & \multicolumn{1}{l}{}               & \textbf{0.93}                       & 0.89                                      \\                         
\bottomrule
\end{tabular}
\end{table}

\section{Real-image interpolation results}
\label{app:ppl}
\begin{figure*}
 \centering

\begin{tabular}{cc}
    \begin{subfigure}[b]{0.45\textwidth}\includegraphics[width=\textwidth]{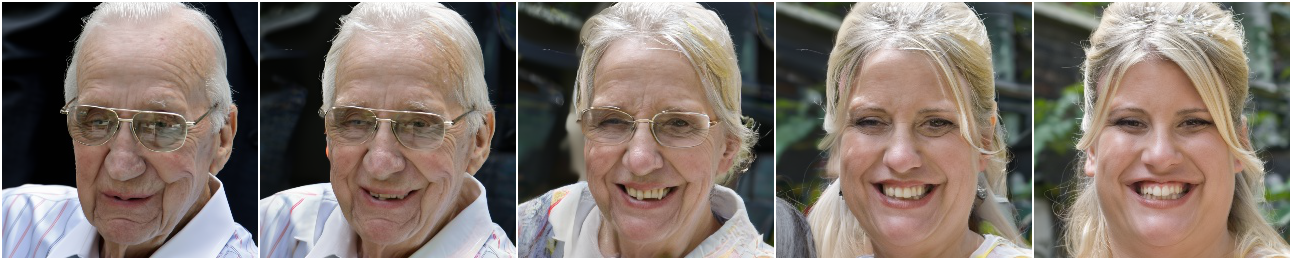}\end{subfigure} &
    \begin{subfigure}[b]{0.45\textwidth}\includegraphics[width=\textwidth]{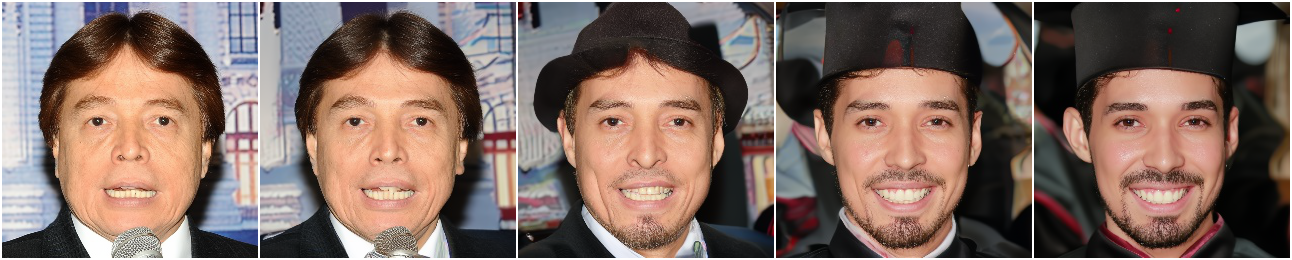}\end{subfigure} \\
    \begin{subfigure}[b]{0.45\textwidth}\includegraphics[width=\textwidth]{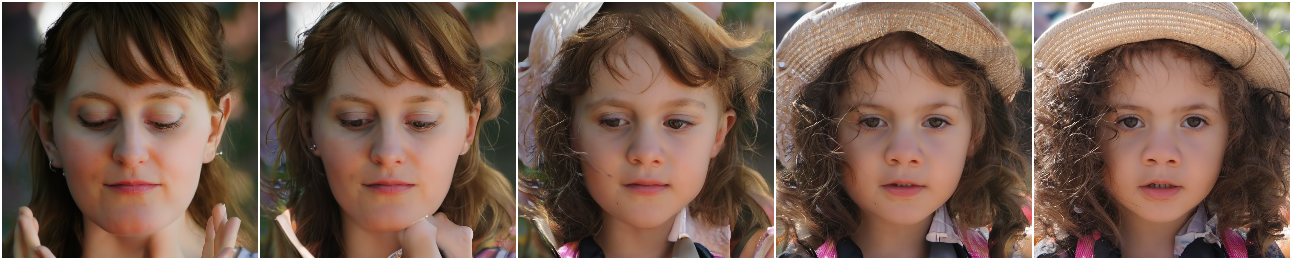}\end{subfigure} &
    \begin{subfigure}[b]{0.45\textwidth}\includegraphics[width=\textwidth]{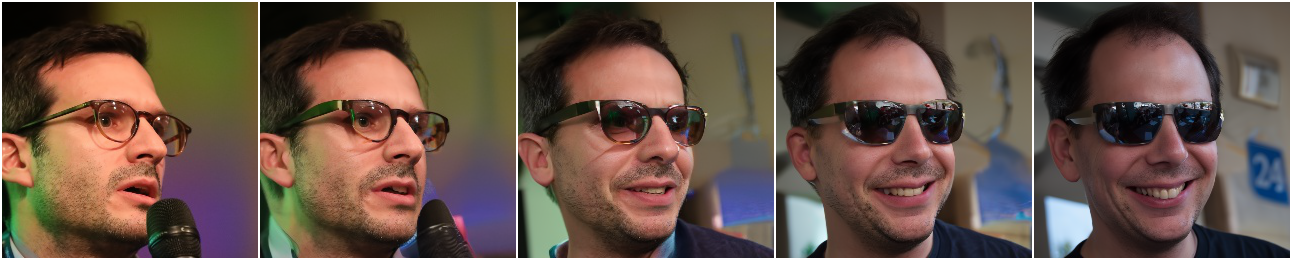}\end{subfigure} 
\end{tabular}
  
  \vspace{-0.2cm}
  \caption{Real-image interpolation on FFHQ dataset \cite{karras_style-based_2019}}
  \label{fig:ffhq_interpo}
  \vspace{-0.2cm}
\end{figure*}

\begin{figure*}
 \centering 
\begin{tabular}{cc}
    \begin{subfigure}[b]{0.45\textwidth}\includegraphics[width=\textwidth]{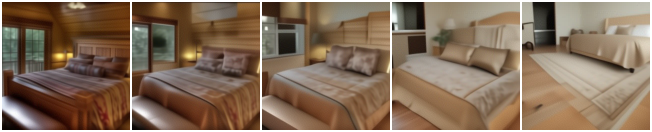}\end{subfigure} &
    \begin{subfigure}[b]{0.45\textwidth}\includegraphics[width=\textwidth]{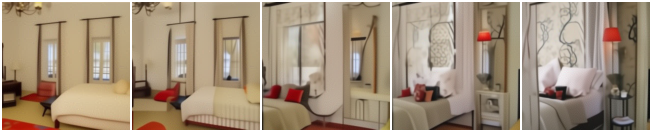}\end{subfigure} \\
    \begin{subfigure}[b]{0.45\textwidth}\includegraphics[width=\textwidth]{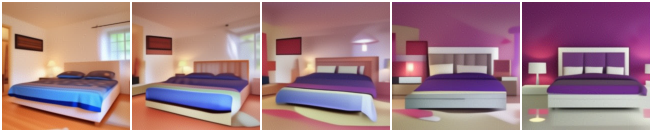}\end{subfigure} &
    \begin{subfigure}[b]{0.45\textwidth}\includegraphics[width=\textwidth]{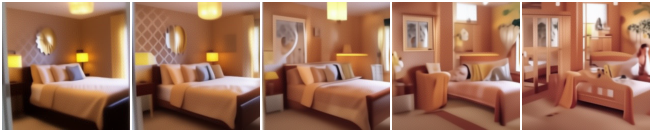}\end{subfigure} 
\end{tabular}
  \vspace{-0.2cm}
  \caption{Real-image interpolation on LSUN bedroom-128 \cite{yu_lsun_2016} }
  \label{fig:bed_interpo}
  \vspace{-0.2cm}
\end{figure*}

\begin{figure*}
 \centering
\begin{tabular}{cc}
    \begin{subfigure}[b]{0.45\textwidth}\includegraphics[width=\textwidth]{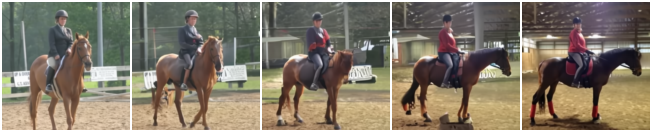}\end{subfigure} &
    \begin{subfigure}[b]{0.45\textwidth}\includegraphics[width=\textwidth]{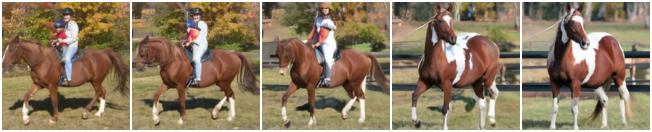}\end{subfigure} \\
    \begin{subfigure}[b]{0.45\textwidth}\includegraphics[width=\textwidth]{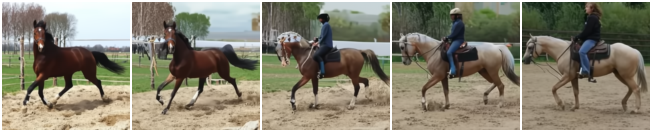}\end{subfigure} &
    \begin{subfigure}[b]{0.45\textwidth}\includegraphics[width=\textwidth]{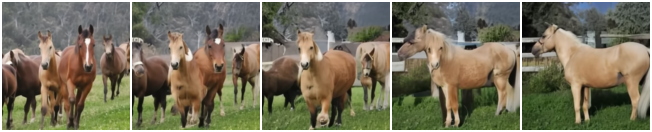}\end{subfigure} 
\end{tabular}
  \vspace{-0.2cm}
  \caption{Real-image interpolation on LSUN horse-128 \cite{yu_lsun_2016}}
  \label{fig:hours_interpo}
  \vspace{-0.2cm}
\end{figure*}

We show interpolation results on real images from FFHQ \cite{karras_style-based_2019} (Figure \ref{fig:ffhq_interpo}), 
LSUN-Bedroom \cite{yu_lsun_2016} (Figure \ref{fig:bed_interpo}) and LSUN-Horse \cite{yu_lsun_2016} (Figure \ref{fig:hours_interpo}). 
Our method can handle challenging morphing between people with and without glasses, bedrooms from different styles and angles, or horses with different body poses.

To quantify the smoothness of the interpolation, we use Perceptual Path Length (PPL) introduced in StyleGAN \cite{karras_style-based_2019}, to measure the perceptual difference in the image as we move along the interpolation path by a small $\epsilon=10^{-4}$ in the latent space. Specifically, we compute the following expectation over multiple sampled pairs of latent codes $(z_1, z_2)$ and $t\in [0, 1]$:

\begin{align} 
\text{PPL} = \mathbb{E} \left[ \frac{1}{\epsilon^2} d(G(\text{slerp}(z_1,z_2;t)),G(\text{slerp}(z_1,z_2;t+\epsilon))\right] 
\end{align}
where $G$ is the decoder, and $d$ computes the perceptual distance based on the VGG16 network. $\text{slerp}(\cdot)$ denotes spherical interpolation. We compute this expected value using 200 samples (400 images) from FFHQ. 
Our method significantly outperforms DDIM in terms of interpolation smoothness as shown below. 

\begin{center}
\begin{tabular}{l|cc}
\toprule
Model   
& DDIM & \textbf{Ours} \\
\midrule
    PPL     
    & 2,634.14   & \textbf{613.73}    \\
\bottomrule
\end{tabular}
\end{center}

\section{Real-image attribute manipulation results}
\label{app:more_attr_manipulation}
We show real-image attribute manipulation results on FFHQ~\cite{karras_style-based_2019} and CelebA-HQ~\cite{karras_progressive_2018} in Figure \ref{fig:attribute} for smiling, wavy hair, aging, and gender change. For more results, please visit {\small \url{ https://Diff-AE.github.io/}}. Our generated results look highly realistic and plausible.

\textbf{FID between the input and its manipulated version.} 
To assess the quality of our manipulated results, we compare their distribution with that of real images with the target positive attribute, such as smiling. Our manipulation is done by moving $\zsem$ linearly along the target direction $\textbf{w}$, found by training a linear classifier (logistic regression) $\mathbf{y} = \mathbf{w}^\top \mathbf{z} + b$ to predict the target attribute using a labeled dataset. The stochastic subcode $\xT$ is kept intact.
Given $\mathbf{z}$, its manipulated version is produced by decoding $\mathbf{z}' = \mathbf{z} + s\mathbf{w}$, 
where $s \in \mathbb{R}$ controls the degree of manipulation. For this experiment, each input image will be manipulated by a different $s_i$ so that the manipulated result reaches the same degree of the target attribute (e.g., similarly big smile) Specifically, we pick $s_i$ so that the logit confidence of its $\mathbf{z}'_i$ equals the median confidence of all real positive images:


\begin{equation}
s_i = \frac{\text{median} - b - \textbf{z}^\top_i \mathbf{w} }{\mathbf{w}^\top \mathbf{w}}    
\end{equation}
In our implementation, we use normalized $\textbf{z}$ instead of $\textbf{z}$ for this operation and unnormalize it before decoding.

In Table \ref{tab:manipulation_fid}, we measure FID scores $\downarrow$ between the manipulated (to be positive) and real positive images, as well as FID scores between real negative and real positive images as baselines for five different attributes from CelebA-HQ \cite{karras_progressive_2018}. While we expect the manipulated images to get closer to the positive images, we also expect them to not deviate too far from the negative as some original content, such as the background, the identity, should be retained. Hence, we also provide FID scores $\downarrow$ between the manipulated images and the real negative images.
Our $\zsem$ manipulated images are closer to the real positive images for 4 out of 5 attributes than those of StyleGAN-$\mathcal{W}$ while better preserving the original contents in all 5 attributes.

\begingroup
\setlength{\tabcolsep}{8pt} 
\begin{table*}
\caption{Image manipulation FID scores $\downarrow$.}
\label{tab:manipulation_fid}
\vspace{-1em}
\center
\begin{tabular}{llccccc}
\toprule
\textbf{Mode}                                     & \textbf{Model}    & \textbf{Male}  & \textbf{Smiling} & \textbf{Wavy Hair} & \textbf{Young} & \textbf{Blond Hair} \\
\midrule
Positive vs negative &                   & 95.82          & 11.15            & 25.04              & 36.75          & 39.65               \\
\hline
\multirow{2}{*}{Manipulated vs. positive} & \textbf{Ours}              & 52.85          & \textbf{9.19}             & \textbf{20.80}              & \textbf{20.68}          & \textbf{33.51}               \\
                                         & StyleGAN-$\mathcal{W}$ & \textbf{42.90} & 18.52   & 27.10     & 31.15 & 33.89      \\
\hline
\multirow{2}{*}{Manipulated vs. negative} & \textbf{Ours}             & \textbf{23.15}          & \textbf{7.25}             & \textbf{4.89}               & \textbf{11.81}          & \textbf{6.79}                \\
                                         & StyleGAN-$\mathcal{W}$ & 66.92 & 22.15   & 20.70     & 31.15 & 27.54      \\   
\bottomrule
\end{tabular}
\end{table*}
\endgroup

\textbf{Identity preservation.} We quantitatively evaluate how well the input's identity is preserved under the manipulation by computing the cosine similarity $\uparrow$ between the ArcFace embeddings~\cite{deng_arcface_2019} of the input and its manipulated version, following \cite{richardson_encoding_2021}.
Table \ref{tab:arcface} shows our scores on CelebA-HQ images of 4 classes  used in Figure \ref{fig:attribute}: Male, Smiling, Wavy Hair, Young. For this experiment, we use the original $\mathcal{W}$ space inversion of StyleGAN that produces the same 512D latent code as our $\zsem$. Their lower scores can be attributed partly to the poor inversion in this space.
\begingroup
\setlength{\tabcolsep}{4pt}
\begin{table}[H]
\caption{Average cosine similarity $\uparrow$ of the ArcFace embeddings~\cite{deng_arcface_2019} of the input and its manipulated version.}
\label{tab:arcface}
\vspace{-1em}
\begin{center}
\begin{tabular}{l|cccc}
\toprule
\textbf{Model} & \textbf{Male}  & \textbf{Smiling} & \textbf{Wavy Hair} & \textbf{Young}   \\
\midrule
    StyleGAN-$\mathcal{W}$     & 0.4174    & 0.7850   & 0.8544  & 0.6955   \\
    \textbf{Ours}         & \textbf{0.6247}    & \textbf{0.8160}   & \textbf{0.9821}  & \textbf{0.8922}   \\
\bottomrule
\end{tabular}
\end{center}
\end{table}
\endgroup

\section{Attribute manipulation comparison to D2C}
\label{app:manipulation}
\begin{figure*}
 \centering
  \includegraphics[width=\textwidth]{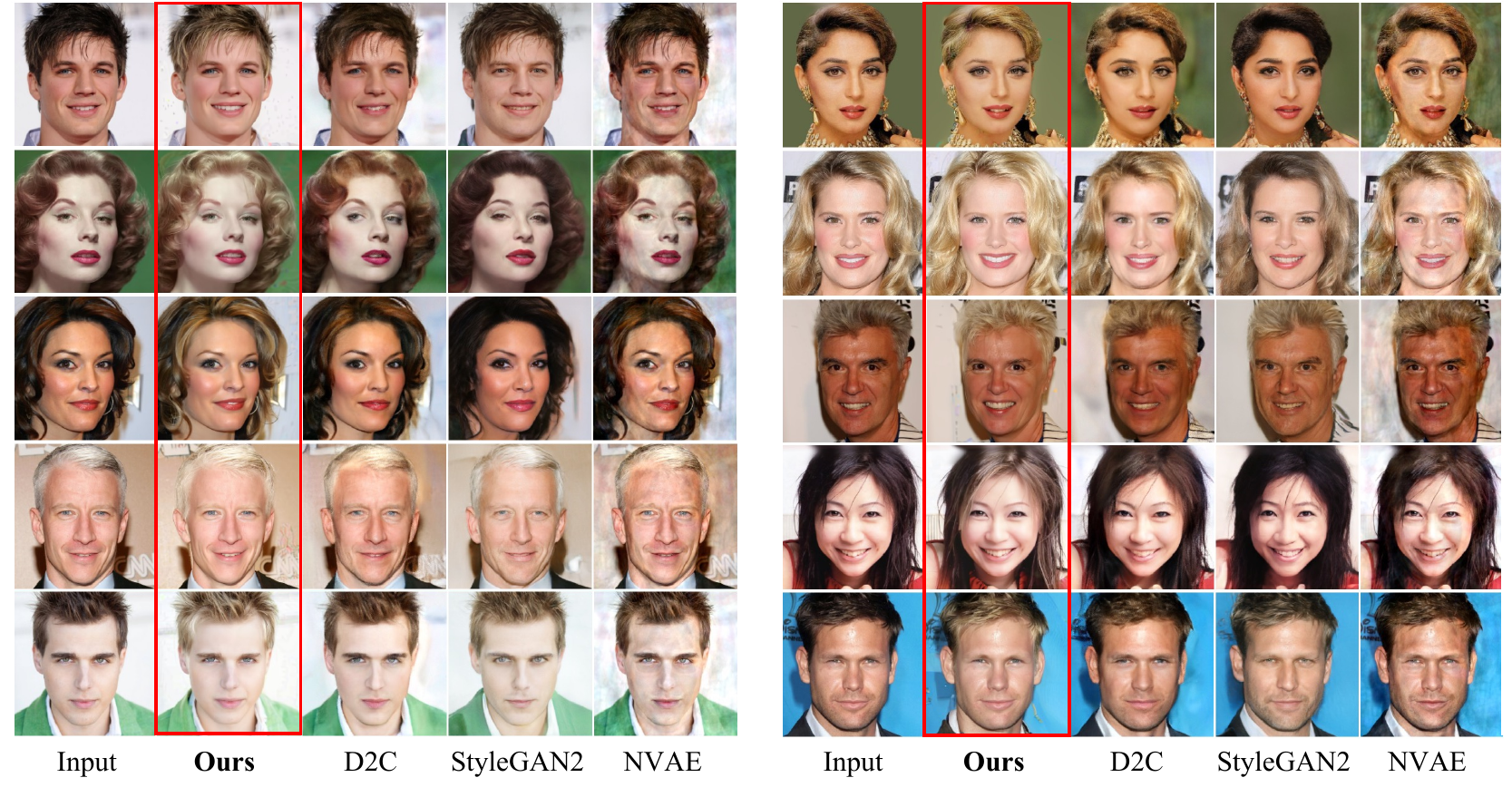}
  \caption{Comparison on attribute manipulation (blond hair) between our method, D2C\cite{sinha_d2c_2021}, StyleGAN2\cite{karras_analyzing_2020}, and NVAE\cite{vahdat_nvae_2020}. 
  }
  \label{fig:manimput}
\end{figure*}

We show a qualitative comparison to D2C \cite{sinha_d2c_2021} on real-image attribute manipulation in Figure \ref{fig:manimput}. These official D2C's results are from {\small \url{ https://d2c-model.github.io/}}. The results of the other baselines are also borrowed from the same website. 

\section{Class-conditional samples}
\begin{figure*}
\begin{center}
    \centering
    \includegraphics[width=1\textwidth]{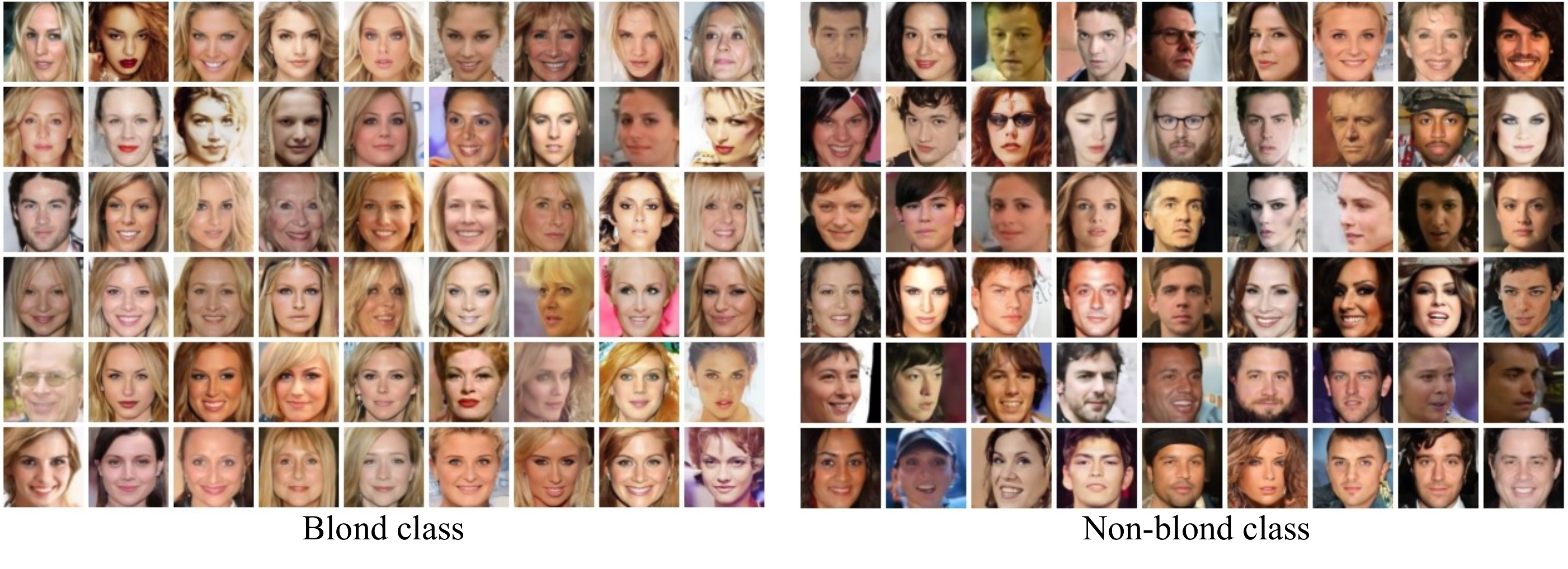}
    \vspace{-0.7cm}
    \caption{Class-conditional generation using 100 positive labeled examples and 10k unlabeled examples on {\it Blond} and {\it Non-blond} from CelebA\cite{karras_progressive_2018}. These results are uncurated. Please see Section \ref{sec:cond_sampling} in the main paper for details.}
    \label{fig:cond_blond}
    \vspace{-0.5cm}
\end{center}
\end{figure*}

\begin{figure*}
\begin{center}
    \centering
    \includegraphics[width=1\textwidth]{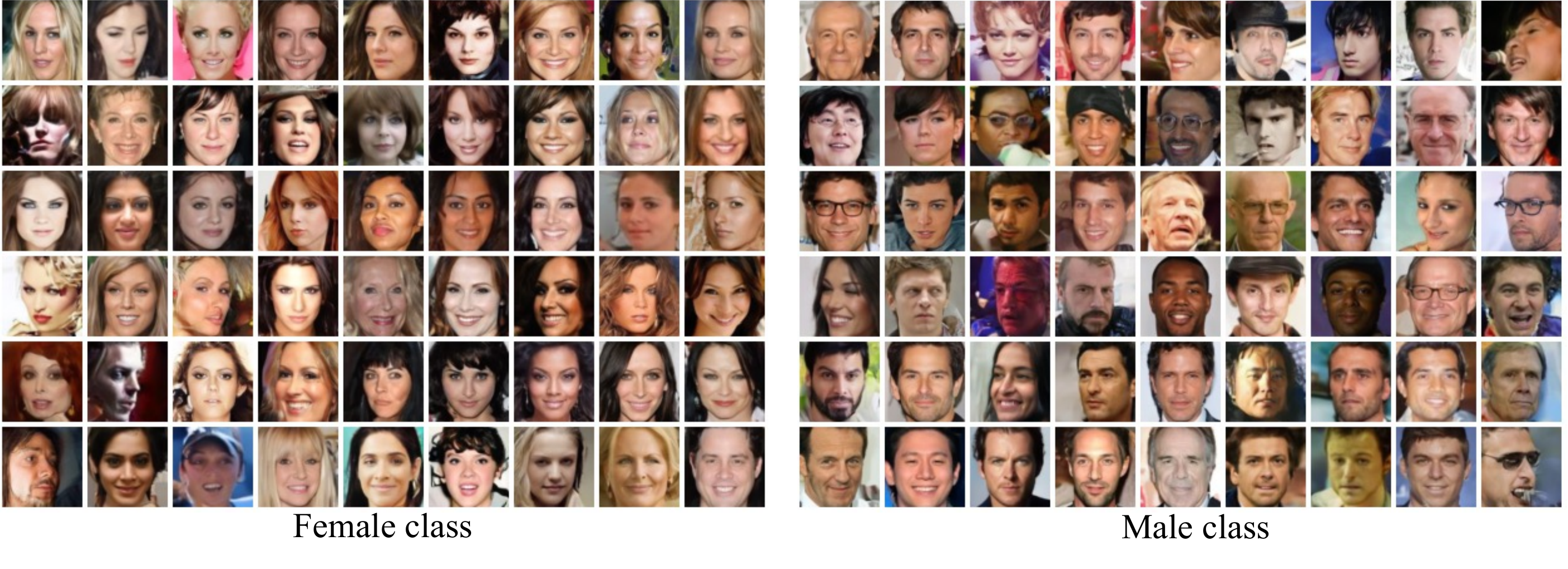}
    \vspace{-0.7cm}
    \caption{Class-conditional generation using 100 positive labeled examples and 10k unlabeled examples on {\it Female} and {\it Male} from CelebA\cite{karras_progressive_2018}. These results are uncurated. Please see Section \ref{sec:cond_sampling} in the main paper for details.}
    \label{fig:cond_female}
    \vspace{-0.5cm}
\end{center}
\end{figure*}

We show our conditional samples of \emph{Blond}, \emph{Non-blond}, \emph{Famale}, and \emph{Male} classes in Figures \ref{fig:cond_blond}, \ref{fig:cond_female}. This is done by training a linear classifier for each attribute using only 100 labeled examples and 10k unlabeled examples, similar to the few-shot experiment done in D2C\cite{sinha_d2c_2021}. The details are in Section \ref{sec:cond_sampling} in the main paper.

\section{Unconditional samples}

\begin{figure*}
\begin{center}
    \centering
    \includegraphics[width=1\textwidth]{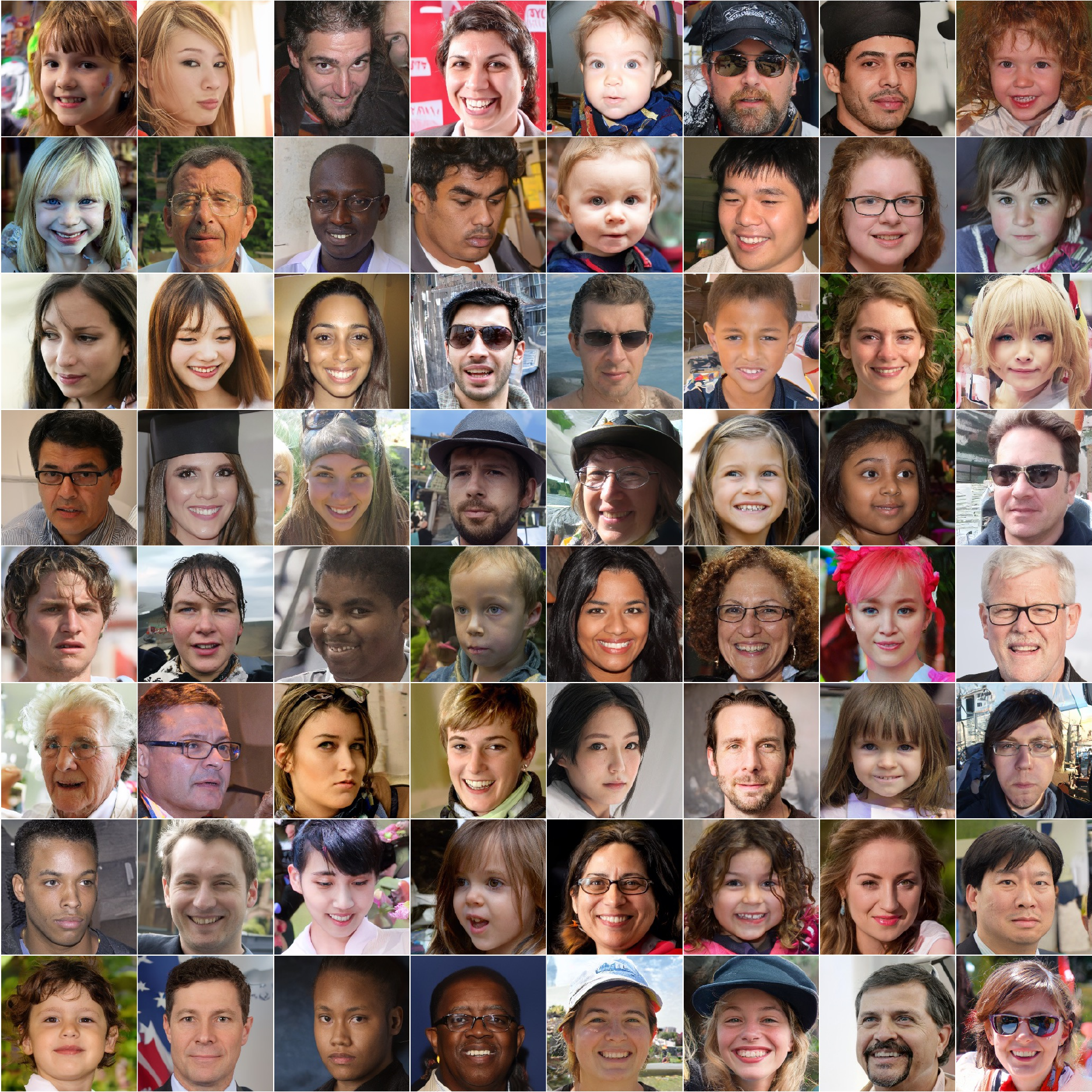}
    \vspace{-0.7cm}
    \caption{Unconditional samples (uncurated) from our diffusion autoencoder and latent DDIM trained on FFHQ-256\cite{karras_style-based_2019}.}
    \label{fig:uncond_face}
    \vspace{-0.5cm}
\end{center}
\end{figure*}

\begin{figure*}
\begin{center}
    \centering
    \includegraphics[width=1\textwidth]{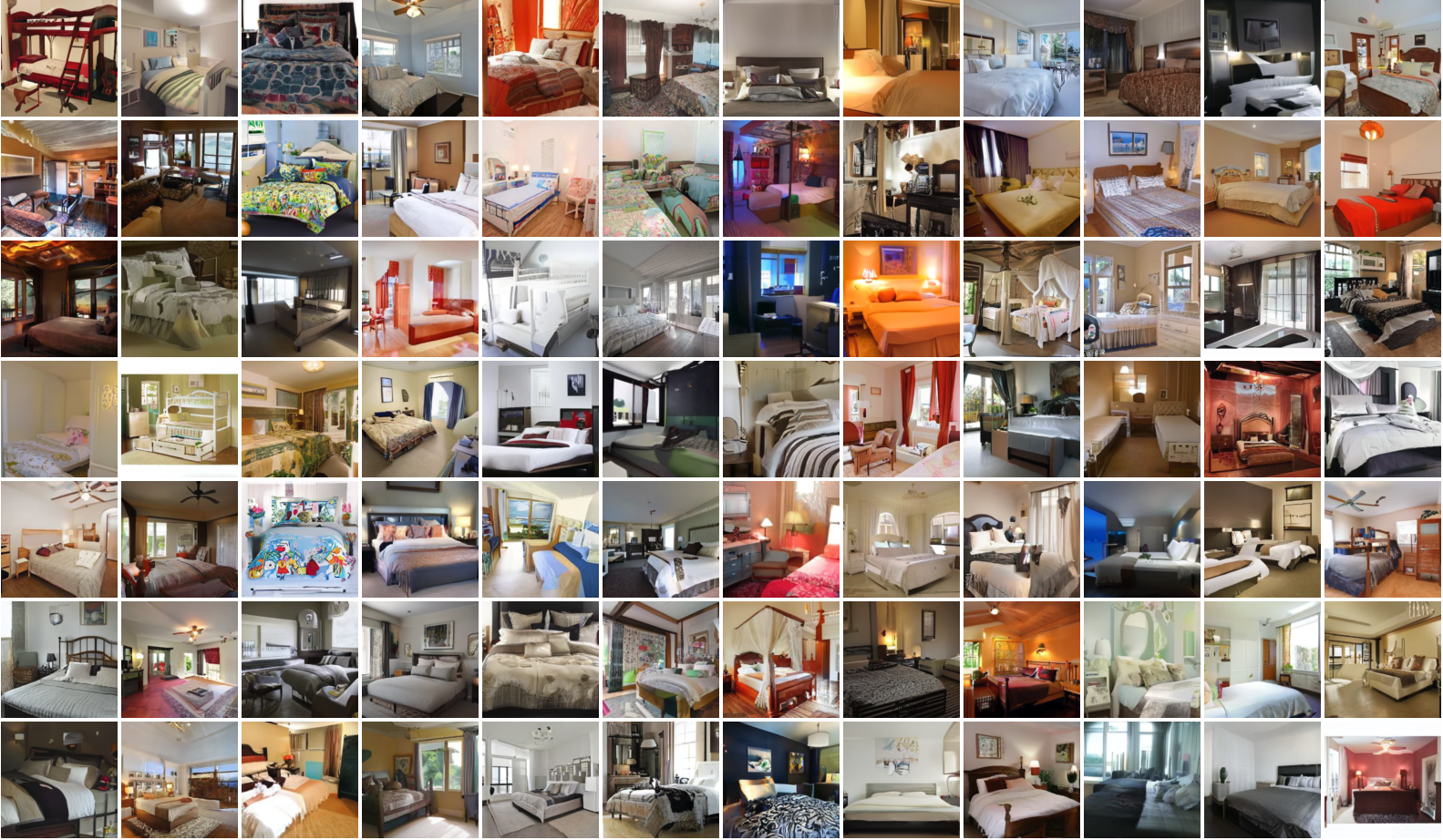}
    \vspace{-0.7cm}
    \caption{Unconditional samples (uncurated) from our diffusion autoencoder and latent DDIM trained on LSUN bedroom-128\cite{yu_lsun_2016}.}
    \label{fig:uncond_bed}
    \vspace{-0.5cm}
\end{center}
\end{figure*}

\begin{figure*}
\begin{center}
    \centering
    \includegraphics[width=1\textwidth]{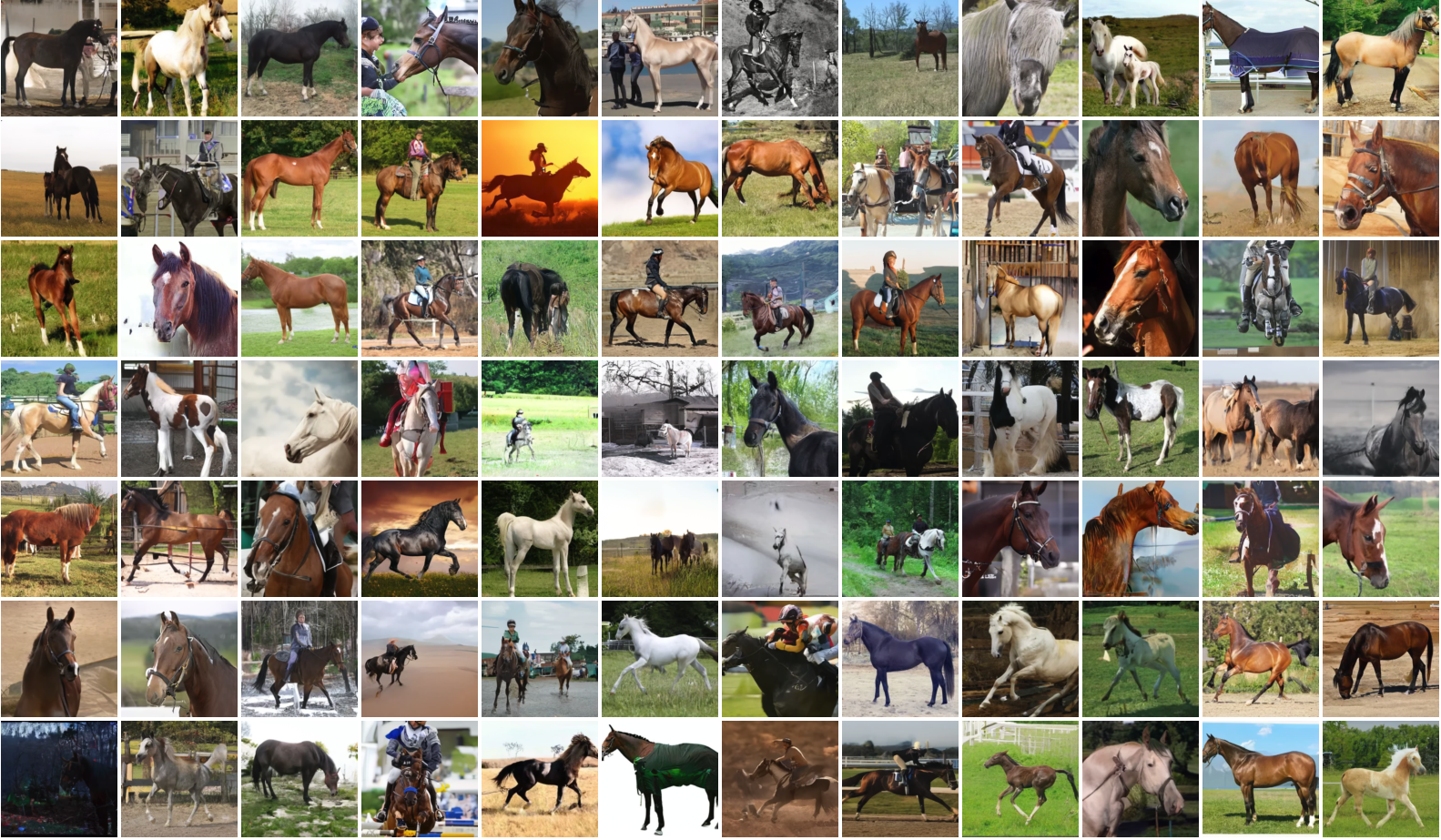}
    \vspace{-0.7cm}
    \caption{Unconditional samples (uncurated) from our diffusion autoencoder and latent DDIM trained on LSUN horse-128\cite{yu_lsun_2016}.}
    \label{fig:uncond_horse}
    \vspace{-0.5cm}
\end{center}
\end{figure*}

We show uncurated unconditional samples from our diffusion autoencoder on FFHQ\cite{karras_style-based_2019}, LSUN-Bed\cite{yu_lsun_2016}, and LSUN-horse\cite{yu_lsun_2016} in Figure \ref{fig:uncond_face}, \ref{fig:uncond_bed}, \ref{fig:uncond_horse}.

\section{Encoding out-of-distribution images}
\begin{figure}[H]
 \centering
 \includegraphics[width=0.48\textwidth]{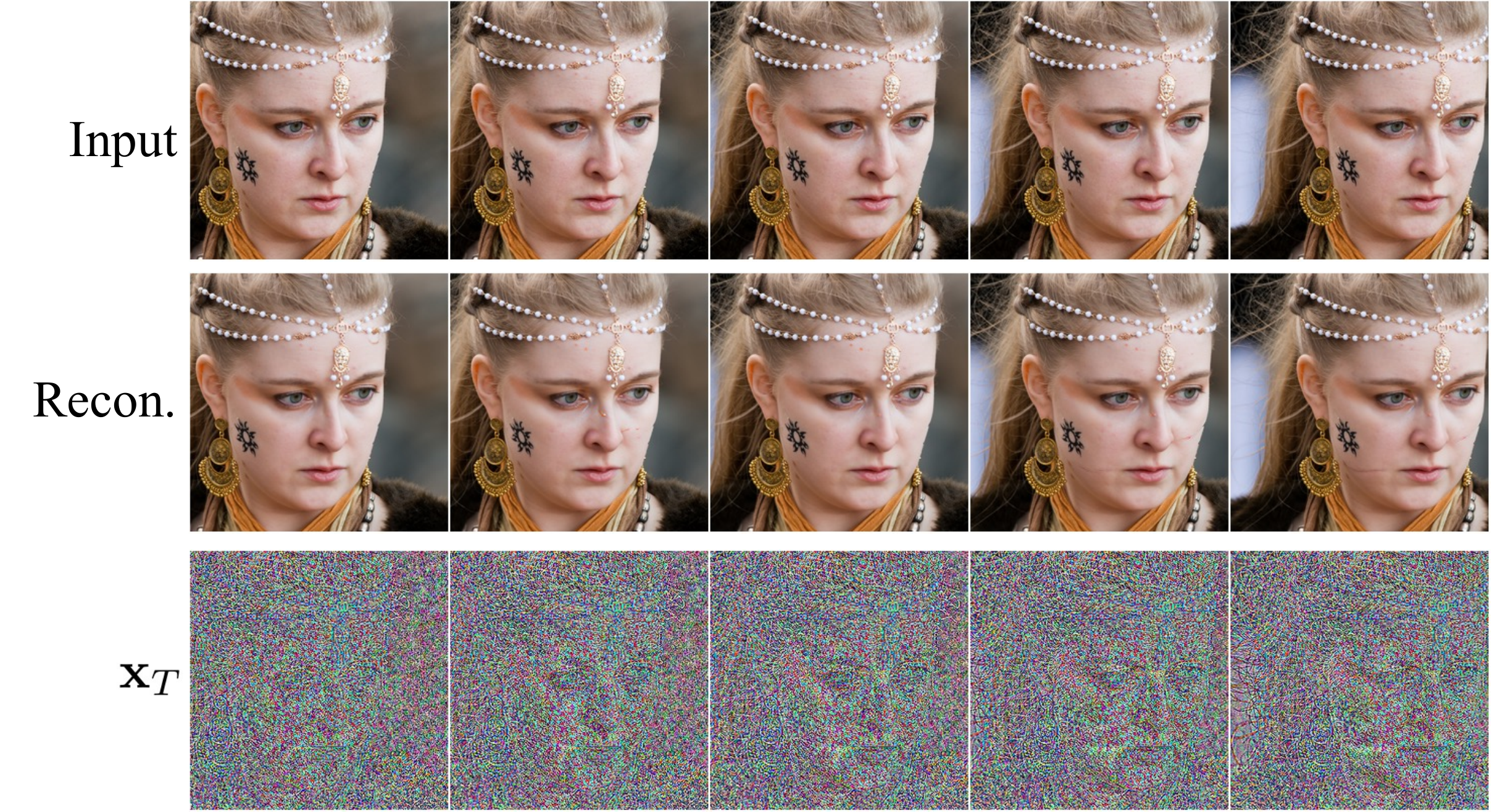}
 \caption{Noise maps $\xT$ when the input face image is shifted to the right to simulate out-of-distribution input image. 
  }
  \label{fig:shift}
\end{figure}

\begin{figure}[H]
 \centering
  \includegraphics[width=0.48\textwidth]{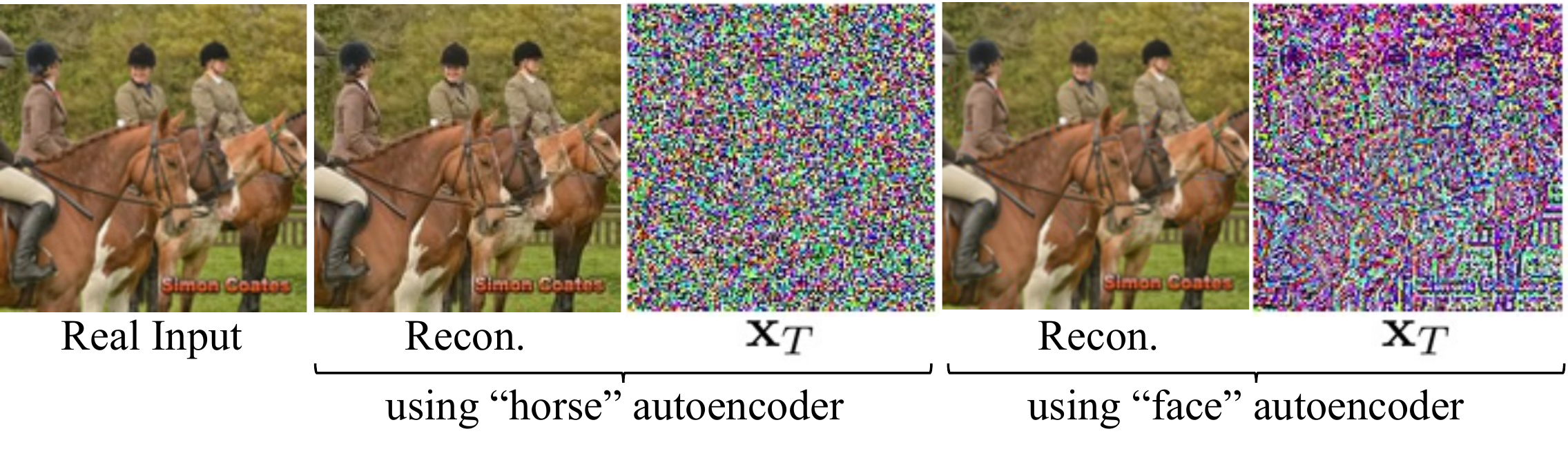}
  \caption{We test how the noise map $\xT$ of a horse image would look if it is encoded by a diffusion autoencoder trained on face images. Both reconstructions look reasonably close to the input image, but $\xT$ from the face autoencoder does not look normally distributed and contains details from the input image.
  }
  \label{fig:horse_fail}
\end{figure}

As discussed in the main paper, when encoding images that are out of the training distribution, our diffusion autoencoders can still reconstruct the images well but the inferred semantic and stochastic subcodes may fall outside the learned distributions. We simulate simple out-of-distribution samples by translating an FFHQ face image in Figure \ref{fig:shift} and by encoding a horse image using our diffusion autoencoder trained on \emph{face} images in Figure \ref{fig:horse_fail}. The reconstruction results still look very close to the input images, but the noise maps $\xT$ show some residual details and do not look normally distributed.

\section{Potential negative impact}
The ability to generate image samples and manipulate attributes of a real image can be used to generate synthetic media, such as deepfakes. We realize the potential negative impact and further conducted a study to determine the difficulty in distinguishing real and synthesized images from our method, as well as discussing some possible directions.

To detect fake images, we train a CNN architecture based on ResNet-50~\cite{he_deep_2016}, which is pretrained on ImageNet~\cite{deng_imagenet_2009}, followed by a linear layer used for classification. Our training dataset consists of ``real'' images from FFHQ256\cite{karras_style-based_2019} and ``fake'' images from either the unconditional sampling experiment (Section \ref{sec:uncond}) or the attribute manipulation experiment (Section \ref{sec:attrib_man}). This dataset contains 20k images: 10K images for each real and fake. The dataset is randomly split into train, test, and validation class-balanced subsets with the ratios of 0.7, 0.2, and 0.1, respectively. The classifier is trained using a binary cross-entropy loss function with an SGD optimizer (learning rate 0.001, momentum 0.9, batchsize 64). Fake detection accuracy is reported here:
\begin{center}
\begin{tabular}{l|lll}
\toprule
\textbf{Method}                 & \textbf{T=100}  & \textbf{T=200}  & \textbf{T=500}   \\
\midrule
Unconditional sampling & 0.9551 & 0.9483 & 0.9313  \\
Attribute manipulation           & 0.9950 & 0.9643 & 0.9213 \\
\bottomrule
\end{tabular}
\label{tab:fake_detection}
\end{center}

The results suggest that even though the generated samples look highly realistic, there could be some certain artifacts that can be easily detected by another neural network. Diffusion-based models also do not have a mechanism to purposely fool a classifier or discriminator like GANs do, and a neural network-based technique is currently found to be $>90\%$ effective at detecting fake images from diffusion models. Note that sampling with higher T leads to samples that are harder to detect. A further study on how easy it is to circumvent detection through adversarial training and an analysis on those giveaway artifacts will be useful for future technical safeguards.  


\begin{figure*}
 \centering
    \includegraphics[width=\textwidth]{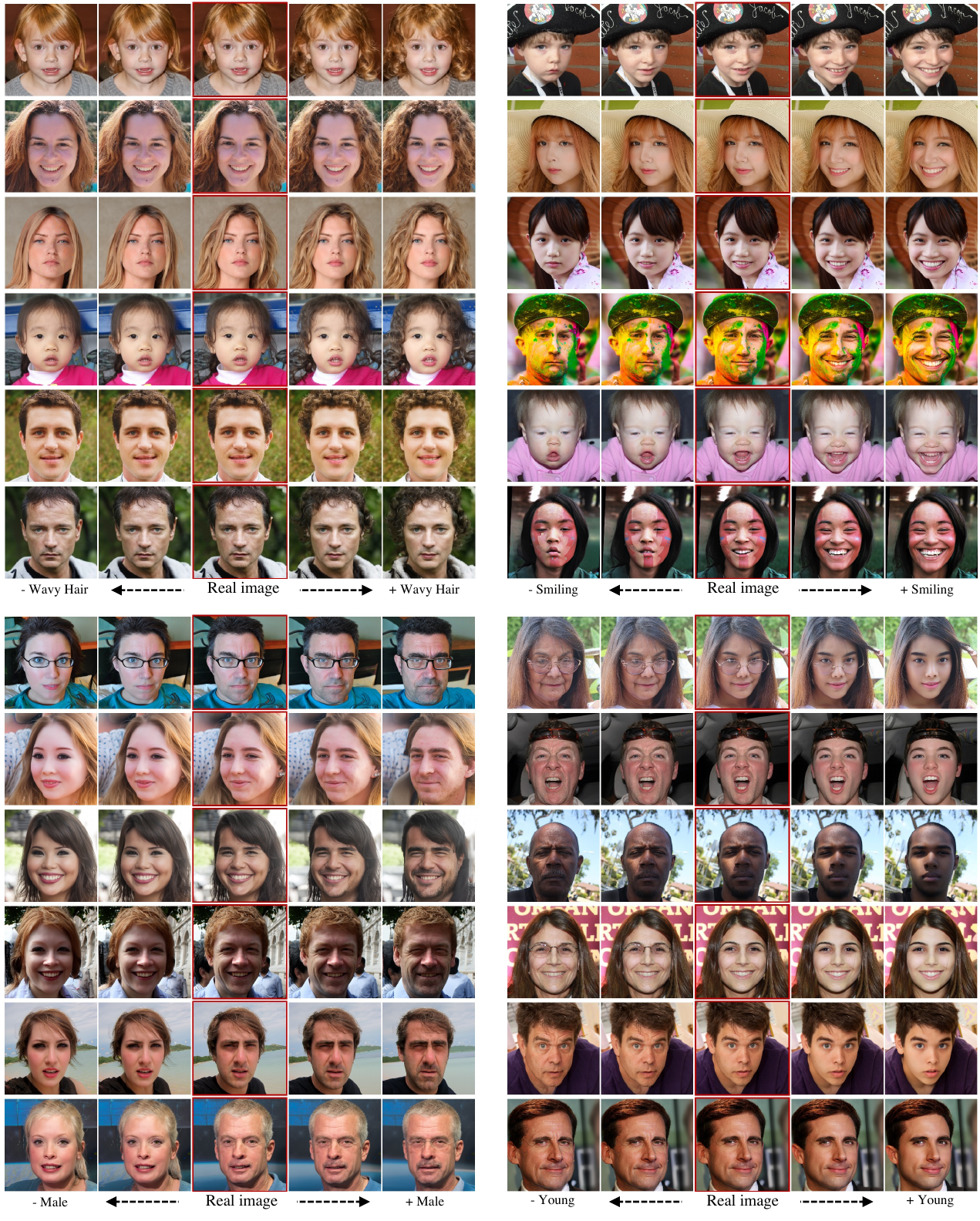}
  \caption{Real-image attribute manipulation for attributes: \emph{Wavy Hair, Smiling, Male, Young}.}
  \label{fig:wavy}
  \vspace{-0.2cm}
\end{figure*}


\end{document}